\documentclass{article}

\usepackage{microtype}
\usepackage{graphicx}
\usepackage{subfig}
\usepackage{booktabs} 
\usepackage{calc}

\usepackage{hyperref}

\usepackage[accepted]{icml2021}

\usepackage{graphicx}
\usepackage{wrapfig}

\usepackage{colortbl}
\usepackage{amsmath}
\usepackage{amssymb}
\usepackage{mathtools}
\usepackage{amsthm}
\usepackage{wrapfig}
\usepackage{cancel}
\usepackage{adjustbox}
\usepackage{multirow}
\usepackage{array}

\usepackage{amsmath,amsfonts,bm}

\def\eqref#1{equation~\ref{#1}}

\def\1{\bm{1}}

\def\vzero{{\bm{0}}}

\def\vtheta{{\bm{\theta}}}
\def\vphi{{\bm{\phi}}}
\def\vepsilon{{\bm{\epsilon}}}
\def\veta{{\bm{\eta}}}
\def\va{{\bm{a}}}
\def\vb{{\bm{b}}}
\def\vc{{\bm{c}}}

\def\vf{{\bm{f}}}
\def\vg{{\bm{g}}}
\def\vh{{\bm{h}}}

\def\vv{{\bm{v}}}
\def\vw{{\bm{w}}}
\def\vx{{\bm{x}}}

\def\vz{{\bm{z}}}

\DeclareMathAlphabet{\mathsfit}{\encodingdefault}{\sfdefault}{m}{sl}
\SetMathAlphabet{\mathsfit}{bold}{\encodingdefault}{\sfdefault}{bx}{n}

\def\gD{{\mathcal{D}}}
\def\gE{{\mathcal{E}}}
\def\gF{{\mathcal{F}}}
\def\gG{{\mathcal{G}}}

\def\gJ{{\mathcal{J}}}

\def\gL{{\mathcal{L}}}

\def\gN{{\mathcal{N}}}

\def\gS{{\mathcal{S}}}

\def\gZ{{\mathcal{Z}}}

\def\sR{{\mathbb{R}}}

\def\sZ{{\mathbb{Z}}}

\newcommand{\E}{\mathbb{E}}

\newcommand{\Cov}{\mathrm{Cov}}

\DeclareMathOperator{\Tr}{Tr}

\newtheorem{theorem}{Theorem}

\newtheorem{myDef}{Definition}
\newtheorem*{remark}{Remark}

\newcommand{\VaES}{\mathrm{VaES}}
\newcommand{\VaGES}{\mathrm{VaGES}}

\newtheorem{innercustomgeneric}{\customgenericname}
\providecommand{\customgenericname}{}
\newcommand{\newcustomtheorem}[2]{%
  \newenvironment{#1}[1]
  {%
   \renewcommand\customgenericname{#2}%
   \renewcommand\theinnercustomgeneric{##1}%
   \innercustomgeneric
  }
  {\endinnercustomgeneric}
}

\newcustomtheorem{customthm}{Theorem}
\newcustomtheorem{customlemma}{Lemma}
\newcustomtheorem{customcorollary}{Corollary}
\newcustomtheorem{customdef}{Definition}

\newlength\celldim \newlength\fontheight \newlength\extraheight
\newcolumntype{x}
{ @{}
>{\centering\arraybackslash}
p{\celldim}
<{\rule[-0.5\extraheight]{0pt}%
{\fontheight + \extraheight}}
@{} }

\icmltitlerunning{Variational (Gradient) Estimate of the Score Function in Energy-based Latent Variable Models}

\begin{document}

\twocolumn[
\icmltitle{Variational (Gradient) Estimate of the Score Function \\ in Energy-based Latent Variable Models}



\icmlsetsymbol{equal}{*}

\begin{icmlauthorlist}
\icmlauthor{Fan Bao}{thu,hua}
\icmlauthor{Kun Xu}{thu}
\icmlauthor{Chongxuan Li}{thu}
\icmlauthor{Lanqing Hong}{hua}
\icmlauthor{Jun Zhu}{thu}
\icmlauthor{Bo Zhang}{thu}
\end{icmlauthorlist}

\icmlaffiliation{thu}{Dept. of Comp. Sci. \& Tech., Institute for AI, THBI Lab, BNRist Center, State Key Lab for Intell. Tech. \& Sys., Tsinghua University, Beijing, China}
\icmlaffiliation{hua}{Huawei Noah's Ark Lab}

\icmlcorrespondingauthor{Jun Zhu}{dcszj@tsinghua.edu.cn}

\icmlkeywords{Machine Learning, ICML}

\vskip 0.3in
]



\printAffiliationsAndNotice{}  

\begin{abstract}
This paper presents new estimates of the \emph{score function} and its gradient with respect to the model parameters in a general energy-based latent variable model (EBLVM).
The score function and its gradient can be expressed as combinations of expectation and covariance terms over the (generally intractable) posterior of the latent variables.
New estimates are obtained by introducing a variational posterior  to approximate the true posterior in these terms.
The variational posterior is trained to minimize a certain divergence (e.g., the KL divergence) between itself and the true posterior. 
Theoretically, the divergence characterizes upper bounds of the bias of the estimates.
In principle, our estimates can be applied to a wide range of objectives, including \emph{kernelized Stein discrepancy} (KSD), \emph{score matching} (SM)-based methods and \emph{exact Fisher divergence} with a minimal model assumption.
In particular, these estimates applied to SM-based methods outperform existing methods in learning EBLVMs on several image datasets.
\end{abstract}

\section{Introduction}
\label{sec:intro}

Energy-based models (EBMs)~\citep{lecun2006tutorial} associate an energy to each configuration of visible variables, which naturally induces a probability distribution by normalizing the exponential negative energy. Such models have found applications in a wide range of areas, such as image synthesis~\cite{du2019implicit,nijkamp2019anatomy}, out-of-distribution detection~\cite{grathwohl2019your,liu2020energy} and controllable generation~\cite{nijkamp2019learning}. Based on EBMs, energy-based latent variable models (EBLVMs)~\cite{swersky2011autoencoders,verteslearning} incorporate latent variables to further improve the expressive power~\cite{salakhutdinov2009deep,bao2020bilevel} and to enable representation learning~\cite{welling2005exponential,salakhutdinov2009deep,srivastava2012multimodal} and conditional sampling (see results in Sec.~\ref{sec:deep_eblvm}).


Notably, the \emph{score function} (SF) of an EBM is defined as the gradient of its log-density with respect to the visible variables, which is independent of the partition function 
and thereby tractable.
Due to its tractability, the SF-based methods are appealing in both learning~\cite{hyvarinen2005estimation,liu2016kernelized} and evaluating~\cite{grathwohl2020learning} EBMs, compared to many other approaches~\cite{hinton2002training,tieleman2008training,gutmann2010noise,salakhutdinov2008learning} (see a comprehensive discussion in Sec.~\ref{sec:related_work}). In particular, \emph{score matching} (SM)~\cite{hyvarinen2005estimation}, which minimizes the expected squared distance between the score function of the data distribution and that of the model distribution, and its variants~\cite{kingma2010regularized,vincent2011connection,saremi2018deep,song2019sliced,li2019annealed,pang2020efficient} have shown promise in learning EBMs. 


Unfortunately, the score function and its gradient with respect to the model parameters in EBLVMs are generally intractable without a strong structural assumption (e.g., the joint distribution of visible and latent variables is in the exponential family~\cite{verteslearning}) and thereby SF-based methods are not directly applicable to general EBLVMs. Recently, bi-level score matching (BiSM)~\cite{bao2020bilevel} applies SM to general EBLVMs by reformulating a SM-based objective as a bilevel optimization problem, which learns a deep EBLVM on natural images and outperforms an EBM of the same size. However, BiSM is not problemless ---  
practically BiSM is optimized by gradient unrolling~\cite{metz2016unrolled} of the lower level optimization, which is time and memory consuming (see a comparison in Sec.~\ref{sec:sm_cmp_grbm}).



In this paper, we present variational estimates of the score function and its gradient with respect to the model parameters in general EBLVMs, referred to as \emph{VaES} and \emph{VaGES} respectively. The score function and its gradient can be expressed as combinations of expectation and covariance terms over the posterior of the latent variables. VaES and VaGES are obtained by introducing a variational posterior, which approximates the true posterior in these terms by minimizing a certain divergence between itself and the true posterior. Theoretically, we show that under some assumptions, the bias introduced by the variational posterior can be bounded by the square root of the KL divergence or the Fisher divergence~\citep{johnson2004information} between the variational posterior and the true one (see Theorem~{\ref{thm:bd_vaes_kl},\ref{thm:bd_vages_kl},\ref{thm:bd_vaes_fisher},\ref{thm:bd_vages_fisher}}).

VaES and VaGES are generally applicable in a wide range of SF-based methods, including \emph{kernelized Stein discrepancy} (KSD)~\cite{liu2016kernelized} (see Sec.~\ref{sec:ksd}) and \emph{score matching} (SM)-based methods~\cite{vincent2011connection,li2019annealed} (see Sec.~\ref{sec:sm}) for learning EBLVMs and estimating the \emph{exact Fisher divergence}~\cite{hu2018stein,grathwohl2020learning} (see Sec.~\ref{sec:eval_fisher}) between the data distribution and the model distribution for evaluating EBLVMs. 
In particular, VaES and VaGES applied to SM-based methods are superior in time and memory consumption compared  with the complex gradient unrolling in the strongest baseline BiSM (see Sec.~\ref{sec:sm_cmp_grbm}) and applicable to learn deep EBLVMs on the MNIST~\cite{lecun2010mnist}, CIFAR10~\cite{krizhevsky2009learning} and CelebA~\cite{liu2015faceattributes} datasets (see Sec.~\ref{sec:deep_eblvm}). We also present latent space interpolation results of deep EBLVMs (see Sec.~\ref{sec:deep_eblvm}), which haven't been investigated in previous EBLVMs to our knowledge.

\section{Method}
As mentioned in Sec.~\ref{sec:intro}, the \emph{score function} has been used in a wide range of methods~\cite{hyvarinen2005estimation,liu2016kernelized,grathwohl2020learning}, while it is generally intractable in EBLVMs without a strong structural assumption. In this paper, we present estimates of the score function and its gradient w.r.t. the model parameters in a general EBLVM. Formally, an EBLVM defines a joint probability distribution over the visible variables $\vv$ and  latent variables $\vh$ as follows
\begin{align}
    p_\vtheta(\vv, \vh) = \tilde{p}_\vtheta(\vv, \vh) / \gZ(\vtheta) = e^{-\gE_\vtheta(\vv, \vh)} / \gZ(\vtheta),
\end{align}
where $\gE_\vtheta(\vv, \vh)$ is the energy function parameterized by $\vtheta$, $\tilde{p}_\vtheta (\vv, \vh) = e^{-\gE_\vtheta(\vv, \vh)}$ is the unnormalized distribution and $\gZ(\vtheta) = \int e^{-\gE_\vtheta(\vv, \vh)} \mathrm{d}\vv \mathrm{d}\vh$ is the partition function. As shown by~\citet{verteslearning}, the score function of an EBLVM can be written as an expectation over the posterior:
\begin{align}
\nabla_{\vv} \log  p_\vtheta(\vv) = \E_{p_\vtheta(\vh|\vv)} \left[ \nabla_{\vv} \log \tilde{p}_\vtheta(\vv, \vh) \right],
\label{eqn:score}
\end{align}
where $p_\vtheta(\vv) = \int p_\vtheta(\vv, \vh) \mathrm{d}\vh$ is the marginal distribution and $p_\vtheta(\vh|\vv)$ is the posterior. We then show that the gradient of the score function w.r.t. the model parameters can be decomposed into an expectation term and a covariance term over the posterior (proof in Appendix~\ref{sec:proof_eqn_grad_score}):
\begin{align}
\label{eqn:grad_score}
& \frac{\partial \nabla_{\vv}\log p_\vtheta(\vv)}{\partial \vtheta} = \underbrace{\E_{p_\vtheta(\vh|\vv)} \left[\frac{\partial \nabla_\vv \log \tilde{p}_\vtheta(\vv, \vh)}{\partial \vtheta}\right]}_{\text{\normalsize{denoted as $e(\vv; \vtheta)$}}} \nonumber \\
& +  \underbrace{\mathrm{Cov}_{p_\vtheta(\vh|\vv)}(\nabla_\vv \log \tilde{p}_\vtheta(\vv, \vh), \nabla_\vtheta \log \tilde{p}_\vtheta(\vv, \vh))}_{\text{\normalsize{denoted as $c(\vv; \vtheta)$}}}, 
\end{align}
where the first term is the expectation of a random matrix $\frac{\partial \nabla_\vv \log \tilde{p}_\vtheta(\vv, \vh)}{\partial \vtheta}$ with $\vh \sim p_\vtheta(\vh|\vv)$ and the second term is the covariance between two random vectors $\nabla_\vv \log \tilde{p}_\vtheta(\vv, \vh)$ and $\nabla_\vtheta \log \tilde{p}_\vtheta(\vv, \vh)$ with $\vh \sim p_\vtheta(\vh|\vv)$.

\subsection{Variational (Gradient) Estimate of the Score Function in EBLVMs}

Eq.~({\ref{eqn:score}) and~(\ref{eqn:grad_score}}) naturally suggest Monte Carlo estimates, while both of them need samples from the posterior $p_\vtheta(\vh|\vv)$, which is generally intractable.
As for approximate inference, we present an amortized variational approach~\cite{kingma2013auto} that considers
\begin{align}
\label{eqn:va_score}
\E_{q_\vphi(\vh|\vv)} \left[\nabla_\vv \log \tilde{p}_\vtheta (\vv, \vh) \right],
\end{align}
and
\begin{align}
\label{eqn:va_grad_score}
&\E_{q_\vphi(\vh|\vv)} \left[\frac{\partial \nabla_\vv \log \tilde{p}_\vtheta(\vv, \vh)}{\partial \vtheta}\right] \nonumber \\
&+ \mathrm{Cov}_{q_\vphi(\vh|\vv)}(\nabla_\vv \log \tilde{p}_\vtheta(\vv, \vh), \nabla_\vtheta \log \tilde{p}_\vtheta(\vv, \vh)),
\end{align}
respectively, where 
$q_\vphi(\vh|\vv)$ is a variational posterior parameterized by $\vphi$. Note that the covariance term in Eq.~{(\ref{eqn:va_grad_score})} is a variational estimate of that in Eq.~{(\ref{eqn:grad_score})}, not derived from taking gradient to Eq.~{(\ref{eqn:va_score})}. We optimize some divergence between $q_\vphi(\vh|\vv)$ and $p_\vtheta(\vh|\vv)$  
\begin{align}
\label{eqn:vi}
\min_\vphi \E_{p_D(\vv)} \left[\gD(q_\vphi(\vh|\vv)||p_\vtheta(\vh|\vv))\right],
\end{align}
where $p_D(\vv)$ denotes the data distribution.
Specifically, if we set $\gD$ in Eq.~(\ref{eqn:vi}) as the KL divergence or the Fisher divergence, then Eq.~(\ref{eqn:vi}) is tractable. For clarity, we refer the readers to Appendix~\ref{sec:tract} for a derivation and a general analysis of the tractability of other divergences.


Below, we consider Monte Carlo estimates based on the variational approximation. According to Eq.~{(\ref{eqn:va_score})}, the variational estimate of the score function (VaES) is
\begin{align}
\label{eqn:vaes}
    \mathrm{VaES}(\vv; \vtheta, \vphi) = \frac{1}{L} \sum\limits_{i=1}^L \nabla_\vv \log \tilde{p}_\vtheta(\vv, \vh_i),
\end{align}
where $\vh_i \stackrel{\text{i.i.d}}{\sim} q_\vphi(\vh|\vv)$ and $L$ is the number of samples from $q_\vphi(\vh|\vv)$. Similarly, according to Eq.~{(\ref{eqn:va_grad_score})}, the variational estimate of the expectation term $e(\vv; \vtheta)$ is
\begin{align}
\label{eqn:sd_est}
    \hat{e}(\vv; \vtheta, \vphi) = \frac{1}{L}\sum\limits_{i=1}^L \frac{\partial \nabla_\vv \log \tilde{p}_\vtheta(\vv, \vh_i)}{\partial \vtheta},
\end{align}

where $\vh_i \stackrel{\text{i.i.d}}{\sim} q_\vphi(\vh|\vv)$. As for the covariance term $c(\vv; \vtheta)$, we estimate it with the sample covariance matrix~\citep{fan2016overview}. According to Eq.~{(\ref{eqn:va_grad_score})}, the variational estimate is
\begin{align}
\label{eqn:cov_est}
    &\hat{c}(\vv;\vtheta, \vphi) = \frac{1}{L-1}\sum\limits_{i=1}^L \nabla_\vv \log \tilde{p}_\vtheta(\vv, \vh_i) \frac{\partial \log \tilde{p}_\vtheta(\vv, \vh_i)}{\partial \vtheta} \nonumber \\
    &- \frac{1}{(L-1)L} \sum_{i=1}^L \nabla_\vv \log \tilde{p}_\vtheta(\vv, \vh_i) \sum_{i=1}^L\frac{ \partial \log \tilde{p}_\vtheta(\vv, \vh_i)}{\partial \vtheta},
\end{align}
where $\vh_i \stackrel{\text{i.i.d}}{\sim} q_\vphi(\vh|\vv)$, $\nabla_\vv \cdot$ outputs column vectors and $\frac{\partial \cdot}{\partial \vtheta}$ outputs row vectors. With Eq.~{(\ref{eqn:sd_est})} and~{(\ref{eqn:cov_est})}, the variational gradient estimate of the score function (VaGES) is
\begin{align}
    \mathrm{VaGES}(\vv; \vtheta, \vphi) = \hat{e}(\vv; \vtheta, \vphi) + \hat{c}(\vv;\vtheta, \vphi).
\end{align}

\textbf{Remark:} Although Eq.~{(\ref{eqn:sd_est})} includes second derivatives, in practice we only need to calculate the product of them and vectors. Similar to the Hessian-vector products~\cite{song2019sliced}, only two backpropagations are required in the calculation (since $\vz^\top \frac{\partial \nabla_\vv \log \tilde{p}_\vtheta(\vv, \vh_i)}{\partial \vtheta} = \frac{\partial \vz^\top \nabla_\vv \log \tilde{p}_\vtheta(\vv, \vh_i)}{\partial \vtheta}$).

\subsection{Bounding the Bias}
\label{sec:bd_bias}

Notice that $\mathrm{VaES}(\vv; \vtheta, \vphi)$ and $\mathrm{VaGES}(\vv; \vtheta, \vphi)$ are actually biased estimates due to the difference between $q_\vphi(\vh|\vv)$ and $p_\vtheta(\vh|\vv)$. Firstly, we show that the bias of $\mathrm{VaES}(\vv; \vtheta, \vphi)$ and $\mathrm{VaGES}(\vv; \vtheta, \vphi)$ can be bounded by the square root of the KL divergence between $q_\vphi(\vh|\vv)$ and $p_\vtheta(\vh|\vv)$ under some assumptions on boundedness, as characterized in Theorem~{\ref{thm:bd_vaes_kl}} and Theorem~{\ref{thm:bd_vages_kl}}. 
\vspace{.2cm}

\begin{theorem}
\label{thm:bd_vaes_kl}
(VaES, KL, proof in Appendix~\ref{sec:proof_bd_kl}) Suppose $\nabla_{\vv} \log \tilde{p}_\vtheta(\vv, \vh)$ is bounded w.r.t. $\vv, \vh$ and $\vtheta$, then the bias of $\mathrm{VaES}(\vv; \vtheta, \vphi)$ can be bounded by the square root of the KL divergence between $q_\vphi(\vh|\vv)$ and $p_\vtheta(\vh|\vv)$ up to multiplying a constant.
\end{theorem}

\begin{theorem}
\label{thm:bd_vages_kl}
(VaGES, KL, proof in Appendix~\ref{sec:proof_bd_kl}) Suppose $\nabla_{\vv} \log \tilde{p}_\vtheta(\vv, \vh)$, $\nabla_{\vtheta} \log \tilde{p}_\vtheta(\vv, \vh)$ and $\frac{\partial \nabla_\vv \log \tilde{p}_\vtheta (\vv, \vh)}{\partial \vtheta}$ are bounded w.r.t. $\vv, \vh$ and $\vtheta$, then the bias of $\mathrm{VaGES}(\vv; \vtheta, \vphi)$ can be bounded by the square root of the KL divergence between $q_\vphi(\vh|\vv)$ and $p_\vtheta(\vh|\vv)$ up to multiplying a constant.
\end{theorem}

\textbf{Remark:} The boundedness assumptions in Theorem~{\ref{thm:bd_vaes_kl},\ref{thm:bd_vages_kl}} are not difficult to satisfy when $\vh$ is discrete. For example, if $\vv$ comes from a compact set (which is naturally satisfied on image data), $\vtheta$ comes from a compact set (which can be satisfied by adding weight decay to $\vtheta$), $\log \tilde{p}_\vtheta (\vv, \vh)$ is second continuously differentiable w.r.t. $\vv$ and $\vtheta$ (which is satisfied in neural networks composed of smooth functions) and $\vh$ comes from a finite set (which is satisfied in most commonly used discrete distributions, e.g., the Bernoulli distribution), then $\nabla_{\vv} \log \tilde{p}_\vtheta(\vv, \vh)$, $\nabla_{\vtheta} \log \tilde{p}_\vtheta(\vv, \vh)$ and $\frac{\partial \nabla_\vv \log \tilde{p}_\vtheta (\vv, \vh)}{\partial \vtheta}$ are bounded w.r.t. $\vv, \vh$ and $\vtheta$ by the extreme value theorem of continuous functions on compact sets.

\begin{figure}[t]
\begin{center}
\subfloat[GRBM, discrete $\vh$]{\includegraphics[width=0.48\linewidth]{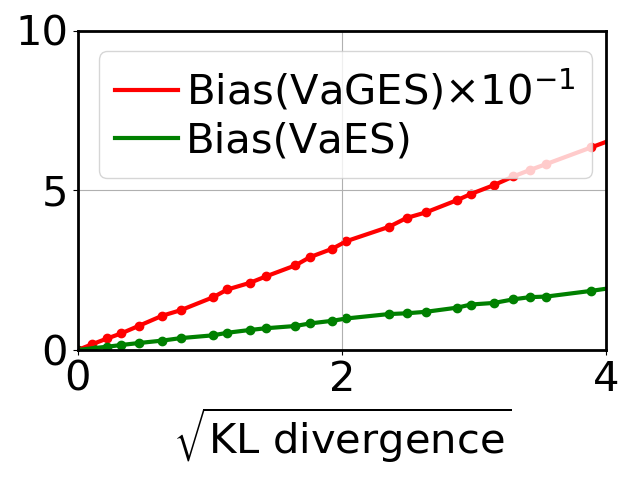}}
\subfloat[GM, continuous $\vh$]{\includegraphics[width=0.48\linewidth]{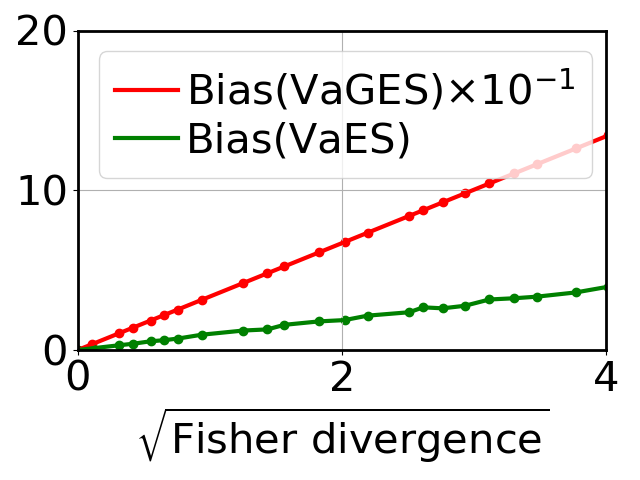}} \\
\caption{The biases of VaES and VaGES v.s. the square root of (a) the KL divergence between $q_\vphi(\vh|\vv)$ and $p_\vtheta(\vh|\vv)$ in a GRBM and (b) the Fisher divergence between $q_\vphi(\vh|\vv)$ and $p_\vtheta(\vh|\vv)$ in a Gaussian model (GM). 
See Appendix~\ref{sec:numer} for experimental details.}
\vspace{-.2cm}
\label{fig:bd_kl_fisher}
\end{center}
\end{figure}

In Fig.~\ref{fig:bd_kl_fisher} (a), we numerically validate Theorem~{\ref{thm:bd_vaes_kl},\ref{thm:bd_vages_kl}} in Gaussian restricted Boltzmann machines (GRBMs)~\citep{hinton2006reducing}. The biases of VaES and VaGES in such models are linearly proportional to the square root of the KL divergence.
Theorem~{\ref{thm:bd_vaes_kl},\ref{thm:bd_vages_kl}} strongly motivate us to set $\gD$ as the KL divergence in Eq.~(\ref{eqn:vi}) when $\vh$ is discrete.

Then, we show that under extra assumptions on the Stein regularity (see Def.~\ref{def:stein_regular}) and boundedness of the Stein factors (see Def.~\ref{def:stein_factor}), the bias of $\mathrm{VaES}(\vv; \vtheta, \vphi)$ and $\mathrm{VaGES}(\vv; \vtheta, \vphi)$ can be bounded by the square root of the Fisher divergence~\citep{johnson2004information} between $q_\vphi(\vh|\vv)$ and $p_\vtheta(\vh|\vv)$, as characterized in Theorem~\ref{thm:bd_vaes_fisher} and Theorem~\ref{thm:bd_vages_fisher}. We first introduce some definitions, which will be used in the theorems.

\begin{myDef}
\label{def:stein_eqn}
{\rm\citep{ley2013stein}} Suppose $p$ is a probability density defined on $\sR^n$ and $f: \sR^n \rightarrow \sR$ is a function, we define $\vg_f^p$ as a solution of the Stein equation $\gS_p \vg = f - \E_p f$, where $\vg: \sR^n \rightarrow \sR^n$ and $\gS_p \vg(\vx) \triangleq \nabla_\vx \log p(\vx)^\top \vg(\vx) + \Tr (\nabla_\vx \vg(\vx))$. (See Appendix~\ref{sec:proof_bd_fisher} for the existence of the solution.)
\end{myDef}

\begin{myDef}
\label{def:stein_regular}
Suppose $p, q$ are probability densities defined on $\sR^n$ and $\vf: \sR^n \rightarrow \sR^m$ is a function, we say $\vf$ satisfies the Stein regular condition w.r.t. $p, q$ iff $\forall i \in \sZ \cap [1, m], \lim\limits_{||\vx|| \rightarrow \infty} q(\vx) \vg_{f_i}^p (\vx) =0 $.
\end{myDef}

\begin{myDef}
\label{def:stein_factor}
{\rm\citep{ley2013stein}} Suppose $p, q$ are probability densities defined on $\sR^n$ and $\vf: \sR^n \rightarrow \sR^m$ is a function satisfying the Stein regular condition w.r.t. $p, q$, we define $\kappa_\vf^{p,q} \triangleq \sqrt{\E_{q(\vx)}\sum\limits_{i=1}^m ||\vg_{f_i}^p (\vx)||_2^2}$, referred to as the Stein factor of $\vf$ w.r.t. $p, q$.
\end{myDef}

\begin{theorem}
\label{thm:bd_vaes_fisher}
(VaES, Fisher, proof in Appendix~\ref{sec:proof_bd_fisher}) Suppose {\rm(1)} $\forall (\vv, \vtheta, \vphi)$, $\nabla_\vv \log \tilde{p}_\vtheta(\vv, \vh)$ as a function of $\vh$ satisfies the Stein regular condition w.r.t. $p_\vtheta(\vh|\vv)$ and $q_\vphi(\vh|\vv)$ and {\rm(2)} the Stein factor of $\nabla_\vv \log \tilde{p}_\vtheta(\vv, \vh)$ as a function of $\vh$ w.r.t. $p_\vtheta(\vh|\vv), q_\vphi(\vh|\vv)$ is bounded w.r.t. $\vv, \vtheta$ and $\vphi$, then the bias of $\mathrm{VaES}(\vv; \vtheta, \vphi)$ can be bounded by the square root of the Fisher divergence between $q_\vphi(\vh|\vv)$ and $p_\vtheta(\vh|\vv)$ up to multiplying a constant.
\end{theorem}

\begin{theorem}
\label{thm:bd_vages_fisher}
(VaGES, Fisher, proof in Appendix~\ref{sec:proof_bd_fisher}) Suppose {\rm(1)} $\forall (\vv, \vtheta, \vphi)$, $\nabla_\vv \log \tilde{p}_\vtheta(\vv, \vh)$, $\nabla_\vtheta \log \tilde{p}_\vtheta(\vv, \vh)$, $\nabla_\vv \log \tilde{p}_\vtheta(\vv, \vh) \frac{\partial \log \tilde{p}_\vtheta(\vv, \vh)}{\partial \vtheta}$ and $\frac{\partial \nabla_\vv \log \tilde{p}_\vtheta(\vv, \vh)}{\partial \vtheta}$ as functions of $\vh$ satisfy the Stein regular condition w.r.t. $p_\vtheta(\vh|\vv)$ and $q_\vphi(\vh|\vv)$ and {\rm(2)} the Stein factors of $\nabla_\vv \log \tilde{p}_\vtheta(\vv, \vh)$, $\nabla_\vtheta \log \tilde{p}_\vtheta(\vv, \vh)$, $\nabla_\vv \log \tilde{p}_\vtheta(\vv, \vh) \frac{\partial \log \tilde{p}_\vtheta(\vv, \vh)}{\partial \vtheta}$ and $\frac{\partial \nabla_\vv \log \tilde{p}_\vtheta(\vv, \vh)}{\partial \vtheta}$ as functions of $\vh$ w.r.t. $p_\vtheta(\vh|\vv), q_\vphi(\vh|\vv)$ are bounded w.r.t. $\vv, \vtheta$ and $\vphi$, {\rm(3)} $\nabla_{\vv} \log \tilde{p}_\vtheta(\vv, \vh)$ and $\nabla_{\vtheta} \log \tilde{p}_\vtheta(\vv, \vh)$ are bounded w.r.t. $\vv, \vh$ and $\vtheta$, then the bias of $\mathrm{VaGES}(\vv; \vtheta, \vphi)$ can be bounded by the square root of the Fisher divergence between $q_\vphi(\vh|\vv)$ and $p_\vtheta(\vh|\vv)$ up to multiplying a constant.
\end{theorem}

Although the boundedness of the Stein factors has only been verified under some simple cases~\citep{ley2013stein}, and hasn't been extended to more complex cases, e.g., when $\tilde{p}_\vtheta(\vv, \vh)$ is parameterized by a neural network, we find it work in practice when choosing $\gD$ in Eq.~{(\ref{eqn:vi})} as the Fisher divergence (see Sec.~\ref{sec:deep_eblvm}) to learn $q_\vphi(\vh|\vv)$. In Fig.~\ref{fig:bd_kl_fisher} (b), we numerically validate Theorem~{\ref{thm:bd_vaes_fisher},\ref{thm:bd_vages_fisher}} in a Gaussian model (GM), whose energy is $\gE_\vtheta(\vv, \vh) = \frac{1}{2\sigma^2} ||\vv - \vb||^2 + \frac{1}{2} ||\vh - \vc||^2 - \vv^\top W \vh$ with $\vtheta=(\sigma, W, \vb, \vc)$. The biases of VaES and VaGES in such models are linearly proportional to the square root of the Fisher divergence.

\subsection{Langevin Dynamics Corrector}
\label{sec:ld_corrector}
Sometimes $p_\vtheta(\vh|\vv)$ can be complex for $q_\vphi(\vh|\vv)$ to approximate, especially in deep models. To improve the inference performance, we can run a few Langevin dynamics steps~\citep{welling2011bayesian} to correct samples from $q_\vphi(\vh|\vv)$, which has been successfully applied to correct the solution of a numerical SDE solver~\citep{song2020score}. Specifically, a Langevin dynamics step updates $\vh$ by
\begin{align}
\label{eqn:ld}
    \vh \leftarrow \vh + \frac{\alpha}{2} \underbrace{\nabla_\vh \log p_\vtheta(\vh|\vv)}_{\substack{\text{\normalsize{\rotatebox{90}{=}}}\\\text{\normalsize{$\nabla_\vh \log \tilde{p}_\vtheta (\vv, \vh)$}}}} + \vepsilon, \ \vepsilon \sim \gN(0, \alpha).
\end{align}

In practice, the standard deviation of $\vepsilon$ is often smaller than $\sqrt{\alpha}$ to allow a faster convergence~\cite{du2019implicit,nijkamp2019anatomy,nijkamp2019learning,grathwohl2019your}. Since Langevin dynamics requires $p_\vtheta(\vh|\vv)$ to be differentiable w.r.t. $\vh$, we only apply the corrector when $\vh$ is continuous.


\vspace{-.1cm}
\section{Learning EBLVMs with KSD}
\label{sec:ksd}

In this section, we show that VaES and VaGES can extend kernelized Stein discrepancy (KSD)~\citep{liu2016kernelized} to learn general EBLVMs. The KSD between the data distribution $p_D(\vv)$ and the model distribution $p_\vtheta(\vv)$ with the kernel $k(\vv, \vv')$ is defined as
\begin{align}
\label{eqn:ksd}
&\quad \mathrm{KSD}(p_D, p_\vtheta) \nonumber \\
&\begin{aligned}[b]
\triangleq \E_{\vv,\vv^\prime\sim p_D} [ & \nabla_\vv \log p_\vtheta(\vv)^\top k(\vv, \vv^\prime) \nabla_{\vv^\prime} \log p_\vtheta(\vv^\prime) \\
& + \nabla_{\vv}\log p_\vtheta(\vv)^\top \nabla_{\vv^\prime} k(\vv, \vv^\prime) \\
& + \nabla_\vv k(\vv, \vv^\prime)^\top \nabla_{\vv^\prime}\log p_\vtheta(\vv^\prime) \\
& + \Tr(\nabla_\vv \nabla_{\vv^\prime} k(\vv, \vv^\prime))],
\end{aligned}
\end{align}
which properly measures the difference between $p_D(\vv)$ and $p_\vtheta(\vv)$ under some mild assumptions~\citep{liu2016kernelized}. To learn an EBLVM, we use the gradient-based optimization to minimize $\mathrm{KSD}(p_D, p_\vtheta)$, where the gradient w.r.t. $\vtheta$ is
\begin{align}
\label{eqn:ksd_grad}
    \frac{\partial \mathrm{KSD}(p_D, p_\vtheta)}{\partial \vtheta} & = 2 \E_{\vv,\vv^\prime\sim p_D} \big[  (k(\vv, \vv^\prime)\nabla_\vv \log p_\vtheta(\vv) \nonumber \\
    & + \nabla_\vv k(\vv, \vv^\prime))^\top\frac{\partial \nabla_{\vv^\prime}\log p_\vtheta(\vv^\prime)}{\partial \vtheta} \big].
\end{align}
 Estimating $\nabla_\vv \log p_\vtheta(\vv)$ and $\frac{\partial \nabla_{\vv^\prime}\log p_\vtheta(\vv^\prime)}{\partial \vtheta}$ with VaES and VaGES respectively, the variational stochastic gradient estimate of KSD (VaGES-KSD) is
\begin{align}
    \frac{2}{M}\sum\limits_{i=1}^M [ & k(\vv_i, \vv_i^\prime) \mathrm{VaES}(\vv_i;\vtheta,\vphi) \nonumber \\
    & + \nabla_\vv k(\vv_i, \vv_i^\prime)]^\top \mathrm{VaGES}(\vv_i';\vtheta,\vphi),
\end{align}
where the union of $\vv_{1:M}$ and $\vv^\prime_{1:M}$ is a mini-batch from the data distribution, and $\mathrm{VaES}(\vv_i;\vtheta,\vphi)$ and $\mathrm{VaGES}(\vv_i';\vtheta,\vphi)$ are independent.

\begin{figure}[t]
\begin{center}
\includegraphics[width=0.85\columnwidth]{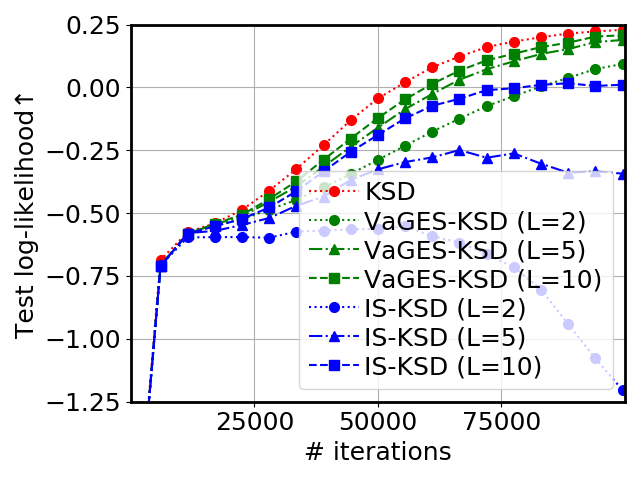}
\vspace{-.4cm}
\caption{Comparison of KSD, VaGES-KSD and IS-KSD on checkerboard. The test log-likelihood is averaged over 10 runs.}
\vspace{-.4cm}
\label{fig:ckbd_ksd}
\end{center}
\end{figure}

\subsection{Experiments}
\label{sec:ksd_exp}

\textbf{Setting.} Since KSD itself suffers from the curse of dimensionality issue~\cite{gong2020sliced}, we illustrate the validity of VaGES-KSD by learning Gaussian restricted Boltzmann machines (GRBMs)~\citep{welling2005exponential,hinton2006reducing} on the 2-D checkerboard dataset, whose distribution is shown in Appendix~\ref{sec:add_ksd}. The energy function of a GRBM is 
\begin{align}
\label{eqn:grbm}
\mathcal{E}_\vtheta(\vv, \vh) = \frac{1}{2\sigma^2} ||\vv - \vb||^2 - \vc^\top  \vh - \vv^\top W \vh,
\end{align}
with learnable parameters $\vtheta = (\sigma, W, \vb, \vc)$. Since GRBM has a tractable posterior, we can directly learn it using KSD (see Eq.~(\ref{eqn:ksd})). We also compare another baseline where $\tilde{p}_\vtheta(\vv)$ is estimated by importance sampling~\cite{mcbook} (i.e., $\tilde{p}_\vtheta(\vv) \approx \frac{1}{L}\sum\limits_{i=1}^L \frac{\tilde{p}_\vtheta(\vv, \vh_i)}{\mathrm{unif}(\vh_i)}, \vh_i \stackrel{\text{i.i.d}}{\sim} \mathrm{unif}(\vh)$, where $\mathrm{unif}$ means the uniform distribution) and we call it IS-KSD. We use the RBF kernel $k(\vv, \vv^\prime) = \exp(-\frac{||\vv-\vv^\prime||_2^2}{2\sigma^2})$ with $\sigma=0.1$. $q_\vphi(\vh|\vv)$ is a Bernoulli distribution parametermized by a fully connected layer with the sigmoid activation and we use the Gumbel-Softmax trick~\citep{jang2016categorical} for reparameterization of $q_\vphi(\vh|\vv)$ with 0.1 as the temperature. $\gD$ in Eq.~{(\ref{eqn:vi})} is the KL divergence. See Appendix~\ref{sec:add_setting_ksd} for more experimental details.

\textbf{Result.} As shown in Fig.~\ref{fig:ckbd_ksd}, VaGES-KSD outperforms IS-KSD on all $L$ (the number of samples from $q_\vphi(\vh|\vv)$) and is comparable to the KSD (the test log-likelihood curves of VaGES-KSD are very close to KSD when $L=5,10$).

\section{Learning EBLVMs with Score Matching}
\label{sec:sm}
In this section, we show that VaES and VaGES can extend two score matching (SM)-based methods~\citep{vincent2011connection,li2019annealed} to learn general EBLVMs.
The first one is the denoising score matching (DSM)~\citep{vincent2011connection}, which minimizes the Fisher divergence $\gD_F$ between a perturbed data distribution $p_{\sigma_0}(\vv)$ and the model distribution $p_\vtheta(\vv)$:
\begin{align}
\label{eqn:dsm}
 &\gJ_{DSM}(\vtheta) \triangleq \gD_F (p_{\sigma_0} || p_\vtheta)  \\
\equiv & \frac{1}{2}\E_{p_D(\vw)p_{\sigma_0}(\vv|\vw)} || \nabla_{\vv} \log  p_\vtheta(\vv) - \nabla_{\vv} \log p_{\sigma_0}(\vv|\vw) ||_2^2, \nonumber
\end{align}
where $p_D(\vw)$ is the data distribution, $p_{\sigma_0}(\vv|\vw)=\gN(\vv|\vw, \sigma_0^2 I)$ is the Gaussian perturbation with a fixed noise level $\sigma_0$, $p_{\sigma_0}(\vv) = \int p_D (\vw) p_{\sigma_0}(\vv|\vw) \mathrm{d} \vw$ is the perturbed data distribution and $\equiv$ means equivalence in parameter optimization. The second one is the multiscale denoising score matching (MDSM)~\citep{li2019annealed}, which uses different noise levels to scale up DSM to high-dimensional data:
\begin{align}
\label{eqn:mdsm}
& \gJ_{MDSM}(\vtheta)  \\
\triangleq &  \frac{1}{2} \E_{p_D(\vw) p(\sigma)  p_\sigma(\vv|\vw)} ||  \nabla_{\vv} \! \log  p_\vtheta(\vv) 
\!-\! \nabla_{\vv} \! \log p_{\sigma_0}(\vv|\vw) ||_2^2, \nonumber
\end{align}
where $p(\sigma)$ is a prior distribution over the flexible noise level $\sigma$ and $\sigma_0$ is a fixed noise level. We extend DSM and MDSM to learn EBLVMs. Firstly we write Eq.~(\ref{eqn:dsm}) and~{(\ref{eqn:mdsm})} in a general form for the convenience of discussion
\begin{align}
    \gJ(\vtheta)\! =\! \frac{1}{2} \E_{p_D(\vw, \vv)} || \nabla_{\vv} \log  p_\vtheta(\vv)\! -\! \nabla_{\vv} \log p_{\sigma_0}(\vv|\vw) ||_2^2 ,
\end{align}
where $p_D(\vw, \vv)$ is the joint distribution of $\vw$ and $\vv$ (specifically, $p_D(\vw, \vv) = p_D(\vw)p_{\sigma_0}(\vv|\vw)$ for Eq.~(\ref{eqn:dsm}) and $p_D(\vw, \vv) = \int_\sigma p_D(\vw)p(\sigma)p_{\sigma}(\vv|\vw) \mathrm{d}\sigma$ for Eq.~(\ref{eqn:mdsm})). We use gradient-based optimization to minimize $\gJ(\vtheta)$ and its gradient w.r.t. $\vtheta$ is
\begin{align}
    \frac{\gJ(\vtheta)}{\partial \vtheta} = & \E_{p_D(\vw, \vv)}  \Big\{  \big[\nabla_{\vv} \log  p_\vtheta(\vv) \nonumber \\
    & -  \nabla_{\vv} \log p_{\sigma_0}(\vv|\vw) \big] \frac{\partial \nabla_{\vv}\log p_\vtheta(\vv)}{\partial \vtheta} \Big\}.
\end{align}
Estimating $\nabla_\vv \log p_\vtheta(\vv)$ and $\frac{\partial \nabla_{\vv}\log p_\vtheta(\vv)}{\partial \vtheta}$ with VaES and VaGES respectively, the variational stochastic gradient estimate of the SM-based methods (VaGES-SM) is
\begin{align}
    \frac{1}{M}\sum\limits_{i=1}^M \big[ & \VaES(\vv_i; \vtheta, \vphi) \nonumber \\
    & - \nabla_{\vv} \log p_{\sigma_0}(\vv_i|\vw_i) \big]  \VaGES(\vv_i; \vtheta, \vphi),
\end{align}
where $(\vw_{1:M}, \vv_{1:M})$ is a mini-batch from $p_D(\vw, \vv)$, and $\mathrm{VaES}(\vv_i;\vtheta,\vphi)$ and $\mathrm{VaGES}(\vv_i;\vtheta,\vphi)$ are independent. We explicitly denote our methods as VaGES-DSM or VaGES-MDSM according to which objective is used.

\subsection{Comparison in GRBMs}
\label{sec:sm_cmp_grbm}
\textbf{Setting.} We firstly consider the GRBM (see Eq.~{(\ref{eqn:grbm})}), which is a good benchmark to compare existing methods. We consider the Frey face dataset, which consists of gray-scaled face images of size $20 \times 28$, and the checkerboard dataset (mentioned in Sec.~\ref{sec:ksd_exp}). $q_\vphi(\vh|\vv)$ is a Bernoulli distribution parametermized by a fully connected layer with the sigmoid activation and we use the Gumbel-Softmax trick~\citep{jang2016categorical} for reparameterization of $q_\vphi(\vh|\vv)$ with 0.1 as the temperature. $\gD$ in Eq.~{(\ref{eqn:vi})} is the KL divergence. See Appendix~\ref{sec:setting_cmp_grbm} for more experimental details.

\begin{figure}[t]
\begin{center}
\subfloat[Test log-likelihood $\uparrow$]{\includegraphics[width=0.48\linewidth]{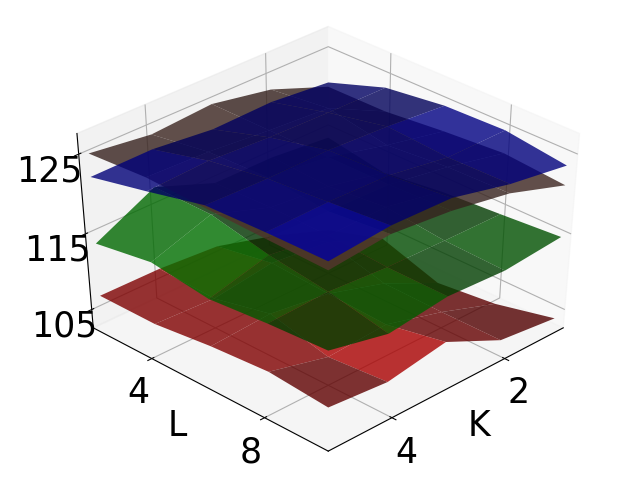}}
\subfloat[Test SM loss $\downarrow$]{\includegraphics[width=0.48\linewidth]{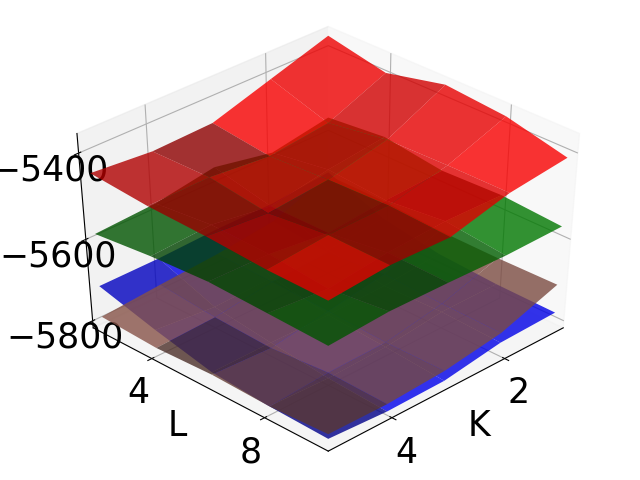}} \\
\subfloat[Time (s) $\downarrow$]{\includegraphics[width=0.48\linewidth]{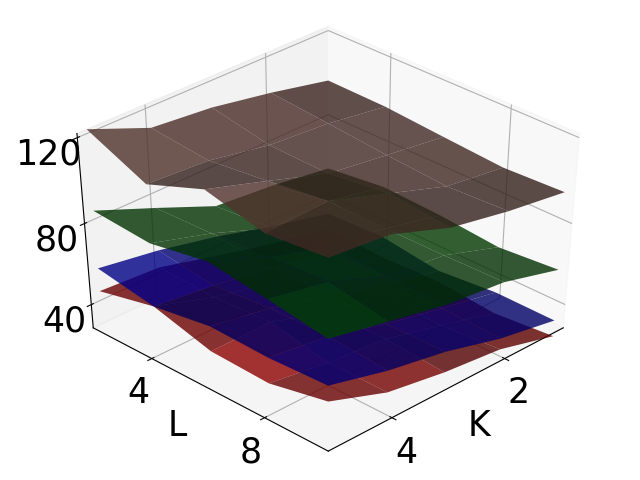}}
\subfloat[Memory (MB) $\downarrow$]{\includegraphics[width=0.48\linewidth]{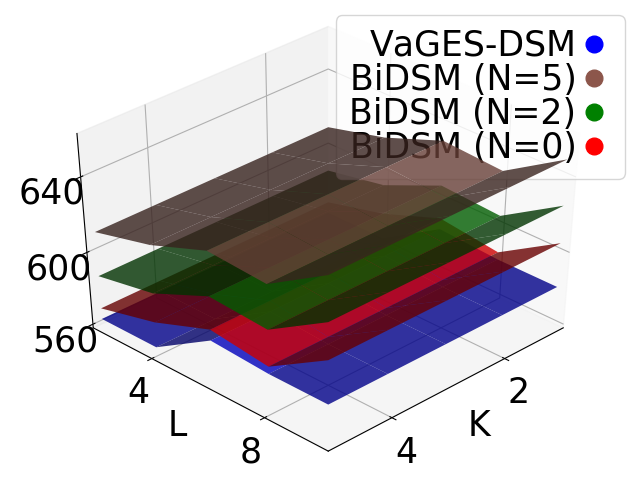}}
\caption{Comparing VaGES-DSM and BiDSM on Frey face. The test SM loss is the Fisher divergence up to a additive constant.}
\vspace{-.4cm}
\label{fig:freyface_cmp}
\end{center}
\end{figure}

\textbf{Comparison with BiSM.} As mentioned in Sec.~\ref{sec:intro}, bi-level score matching (BiSM)~\citep{bao2020bilevel} learns general EBLVMs by reformulating Eq.~{(\ref{eqn:dsm})} and~{(\ref{eqn:mdsm})} as bi-level optimization problems, referred to as BiDSM and BiMDSM respectively, which serves as the most direct baseline (see Appendix~\ref{sec:intro_bism} for an introduction to BiSM). We make a comprehensive comparison between VaGES-SM and BiSM on the Frey face dataset, which explores all hyperparameters involved in these two methods. Specifically, both methods sample from a variational posterior and optimize the variational posterior, so they share two hyperparameters $L$ (the number of $\vh$ sampled from the variational posterior) and $K$ (the number of times updating the variational posterior on each mini-batch). Besides, BiSM has an extra hyperparameter $N$, which specifies the number of gradient unrolling steps. In Fig.~\ref{fig:freyface_cmp}, we plot the test log-likelihood, the test score matching (SM) loss~\cite{hyvarinen2005estimation}, which is the Fisher divergence up to an additive constant, the time and memory consumption by grid search across $L$, $K$ and $N$. VaGES-DSM takes about $ 40\%$ time and $90\%$ memory on average to achieve a similar performance with BiDSM ($N$=5). VaGES-DSM outperforms BiDSM ($N$=0,2), and meanwhile it takes less time than BiDSM ($N$=2) and the least memory.

\textbf{Comparison with other methods.} For completeness, we also compare with DSM~\cite{vincent2011connection}, CD-based methods~\cite{hinton2002training,tieleman2008training} and noise contrastive estimation (NCE)-based methods~\cite{gutmann2010noise,pmlr-v89-rhodes19a} on the checkerboard dataset. VaGES-DSM is competitive to the baselines (e.g., CD and DSM) that leverage the tractability of the posterior in GRBM and outperforms the variational NCE~\cite{pmlr-v89-rhodes19a}. See results in Appendix~\ref{sec:add_cmp_grbm}.

\subsection{Learning Deep EBLVMs} 
\label{sec:deep_eblvm}

\textbf{Setting.} Then we show that VaGES-SM can scale up to learn general deep EBLVMs and generate natural images. Following BiSM~\cite{bao2020bilevel}, we consider the deep EBLVM with the following energy function
\begin{align}
\label{eqn:deep_eblvm}
\mathcal{E}_\vtheta(\vv, \vh) = g_3(g_2(g_1(\vv; \vtheta_1), \vh); \vtheta_2),
\end{align}
where $\vh$ is continuous, $\vtheta = (\vtheta_1, \vtheta_2)$, $g_1$ a vector-valued neural network that outputs a feature sharing the same dimension with $\vh$, $g_2$ is an additive coupling layer~\cite{dinh2014nice} and $g_3$ is a scalar-valued neural network. We consider the MNIST dataset~\cite{lecun2010mnist}, which consists of gray-scaled hand-written digits of size $28\times28$, the CIFAR10 dataset~\cite{krizhevsky2009learning}, which consists of color natural images of size $32\times 32$ and the CelebA dataset~\cite{liu2015faceattributes}, which consists of color face images. We resize CelebA to $64\times 64$. $q_\vphi(\vh|\vv)$ is a Gaussian distribution parameterized by a 3-layer convolutional neural network. $\gD$ in Eq.~{(\ref{eqn:vi})} is the Fisher divergence. We use the Langevin dynamics corrector (see Sec.~\ref{sec:ld_corrector}) to correct samples from $q_\vphi(\vh|\vv)$. As for the step size $\alpha$ in Langevin dynamics, we empirically find that the optimal one is approximately proportional to the dimension of $\vh$ (see Appendix~\ref{sec:setting_learn_eblvm}), perhaps because Langevin dynamics converges to its stationary distribution slower when $\vh$ has a higher dimension and thereby requires a larger step size, so we fix the ratio of the step size $\alpha$ in Langevin dynamics to the dimension of $\vh$ on one dataset and the ratio is $2 \times 10^{-4}$ on MNIST and CIFAR10 and $10^{-5}$ on CelebA. The standard deviation of the noise $\vepsilon$ in Langevin dynamics is $10^{-4}$. See Appendix~\ref{sec:setting_learn_eblvm} for more experimental details, e.g., how hyperparameters are selected and time consumption.

\textbf{Sample quality.} Since the EBLVM defined by Eq.~{(\ref{eqn:deep_eblvm})} has an intractable posterior, BiSM is the only applicable baseline mentioned in Section~\ref{sec:sm_cmp_grbm}.  We quantitatively evaluate the sample quality with the FID score~\cite{heusel2017gans} in Tab.~\ref{tab:fid} and VaGES-MDSM achieves comparable performance to BiMDSM. Since there are relatively few baselines of learning deep EBLVMs, we also compare with other baselines involving learning EBMs in Tab.~\ref{tab:fid}. We mention that VAE-EBLVM~\cite{han2020joint} and CoopNets~\cite{xie2018cooperative} report FID results on a subset of CelebA and on a different resolution of CelebA respectively. For fairness, we don't include these results on CelebA. We provide image samples and Inception Score results in Appendix~\ref{sec:add_learn_eblvm}.
\begin{table}[t]
\centering
\caption{FID on CIFAR10 and CelebA ($64\times 64$). $^\dagger$ Averaged by 5 runs. $^\ddagger$ Since BiSM doesn't report a FID on CelebA, the value is evaluated in our reproduction.}
\label{tab:fid}
\vspace{.2cm}
\subfloat[CIFAR10]{\begin{tabular}{cc}
\toprule
    Methods & FID $\downarrow$ \\
    \midrule
      Flow-CE~\cite{gao2020flow} & 37.30  \\
     VAE-EBLVM~\cite{han2020joint} & 30.1 \\
     CoopNets~\cite{xie2018cooperative} & 33.61 \\
     EBM~\cite{du2019implicit} & 38.2 \\
    MDSM~\cite{li2019annealed} & 31.7 \\
    BiMDSM~\cite{bao2020bilevel}  & 29.43 $\pm$ 2.76$^\dagger$ \\
    VaGES-MDSM (ours) & \textbf{28.93} $\pm$ 1.91$^\dagger$ \\
\bottomrule
\end{tabular}} \\
\subfloat[CelebA]{\begin{tabular}{cc}
\toprule
   Methods & FID $\downarrow$ \\
    \midrule
    BiMDSM~\cite{bao2020bilevel} & 32.43$^\ddagger$ \\
    VaGES-MDSM (ours) & \textbf{31.53} \\
\bottomrule
\end{tabular}}
\vspace{-.6cm}
\end{table}


\begin{figure*}[t]
\begin{center}
\subfloat[MNIST]{\includegraphics[width=0.3\linewidth]{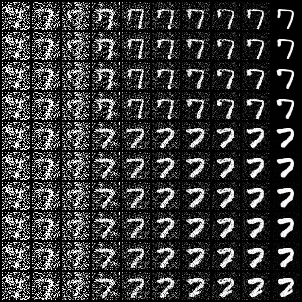}}\quad
\subfloat[CIFAR10]{\includegraphics[width=0.3\linewidth]{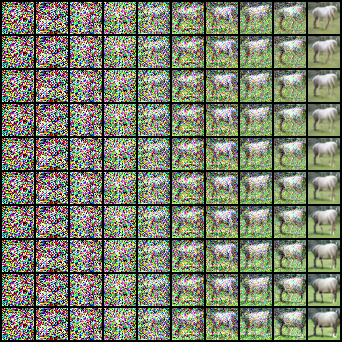}}\quad
\subfloat[CelebA]{\includegraphics[width=0.3\linewidth]{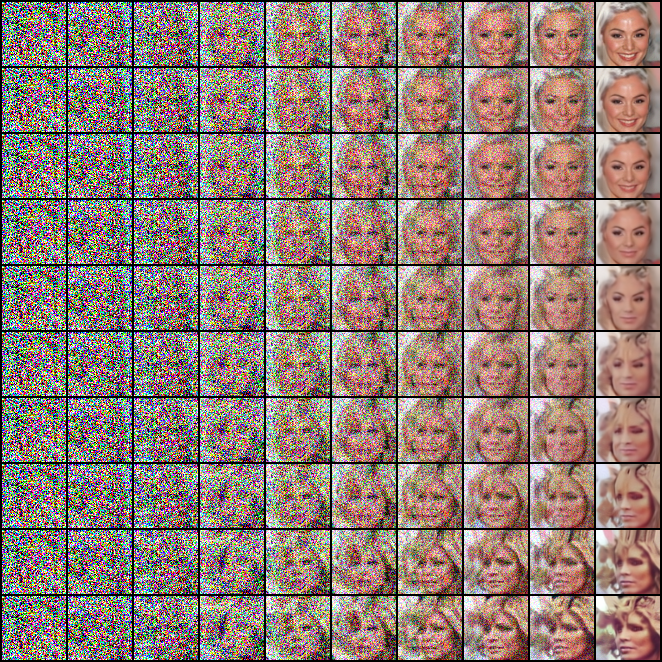}}
\vspace{-.2cm}
\caption{Interpolation of annealed Langevin dynamics trajectories in the latent space. Each row displays a trajectory of $p_\vtheta(\vv|\vh)$ for a fixed latent variable $\vh$ and $\vh$ changes smoothly between rows. Different trajectories share the same initial point.}
\vspace{-.4cm}
\label{fig:trajectory}
\end{center}
\end{figure*}

\textbf{Interpolation in the latent space.} 
As argued in prior work~\cite{bao2020bilevel}, the conditional distribution $p_\vtheta(\vv|\vh)$ of a deep EBLVM can be multimodal, so we want to explore how a local modal of $p_\vtheta(\vv|\vh)$ evolves as $\vh$ changes. Since $\vv$ is of high dimension, we are not able to plot the landscape of $p_\vtheta(\vv|\vh)$. Alternatively, we study how a fixed particle moves in different landscapes according to the annealed Langevin dynamics~\cite{li2019annealed} by interpolating $\vh$. Specifically, we select two points in the latent space and interpolate linearly between them to get a class of conditional distributions $\{p_\vtheta(\vv|\vh_i)\}_{i=1}^T$. Then we fix an initial point $\vv_0$ and run annealed Langevin dynamics on these conditional distributions with a shared noise. These annealed Langevin dynamics trajectories are shown in Fig.~\ref{fig:trajectory}. Starting from the same initial noise, the trajectory evolves smoothly as $\vh$ changes smoothly. More interpolation results can be found in Appendix~\ref{sec:add_learn_eblvm}.

\textbf{Sensitivity analysis.} We also study how hyperparameters influence the performances of VaGES-SM in deep EBLVMs, including the dimension of $\vh$, the number of convolutional layers, the number of $\vh$ sampled from $q_\vphi(\vh|\vv)$, the number of times of updating $\vphi$, the number of Langevin dynamics steps, the noise level in Langevin dynamics and which divergence to use in Eq.~{(\ref{eqn:vi})}. These results can be found in Appendix~\ref{sec:add_learn_eblvm}. In brief, increasing the number of convolutional layers will improve the performance, while the dimension of $\vh$, the number of $\vh$ sampled from $q_\vphi(\vh|\vv)$, the noise level in Langevin dynamics and using the KL divergence instead of the Fisher divergence in Eq.~{(\ref{eqn:vi})} don't affect the result very much. Besides, setting both the number of times updating $\vphi$ and the number of Langevin dynamics steps to 5 is enough for a stable training process.

\section{Evaluating EBLVMs with Exact Fisher Divergence}
\label{sec:eval_fisher}

In this setting we are given an EBLVM and a set of samples $\{\vv_i\}_{i=1}^n$ from the data distribution $p_D(\vv)$. We want to measure how well the model distribution $p_\vtheta(\vv)$ approximates $p_D(\vv)$, which needs an absolute value representing the difference between them.
This task is more difficult than model comparison, which only needs relative values (e.g., log-likelihood) to compare between different models.
Specifically, we consider the Fisher divergence of the maximum form~\citep{hu2018stein,grathwohl2020learning} to measure how well $p_\vtheta(\vv)$ approximates $p_D(\vv)$:
\begin{align}
\label{eqn:fisher_m}
    \gD_F^m(p_D || p_\vtheta) \nonumber  \triangleq  & \max\limits_{\vf \in \gF} \E_{p_D(\vv)} \E_{p(\vepsilon)} \big[ \nabla_\vv \log p_\vtheta(\vv)^\top \vf(\vv) \nonumber \\
    &   + \vepsilon^\top \nabla_\vv \vf(\vv) \vepsilon - \frac{1}{2}||\vf(\vv)||_2^2 \big],
\end{align}
where $\gF$ is the set of functions $\vf: \sR^d \rightarrow \sR^d$ satisfying $\lim_{||\vv|| \rightarrow \infty} p_D(\vv) \vf(\vv) = \vzero$ and $p(\vepsilon)$ is a noise distribution (e.g., Gaussian distribution) introduced for computation efficiency~\citep{grathwohl2020learning}. Such a form can also be understood as the Stein discrepancy~\citep{gorham2017measuring} with $\gL_2$ constraint on the function space $\gF$ and has been applied to evaluate GRBMs~\citep{grathwohl2020learning}. Under some mild assumptions (see Appendix~\ref{sec:consist}), Eq.~{(\ref{eqn:fisher_m})} is equal to the exact Fisher divergence. In practice, $\gF$ is approximated by a neural network $\{ \vf_\veta: \veta \in H \}$, where $\veta$ is the parameter and $H$ is the parameter space. We optimize the right hand side of Eq.~{(\ref{eqn:fisher_m})} and the gradient w.r.t. $\veta$ is
\begin{align}
    \E_{p_D(\vv)} \E_{p(\vepsilon)} \big[ & \nabla_\vv \log p_\vtheta(\vv)^\top \frac{\partial\vf_\veta(\vv)}{\partial \veta} \nonumber \\
    & + \frac{\partial \vepsilon^\top \nabla_\vv \vf_\veta(\vv) \vepsilon}{\partial \veta} - \frac{1}{2}\frac{\partial ||\vf_\veta(\vv)||_2^2}{\partial \veta} \big].
\end{align}
Estimating $\nabla_\vv \log p_\vtheta(\vv)$ with VaES, the variational stochastic gradient estimate is
\begin{align}
    \frac{1}{M}\sum_{i=1}^M \big[ & \VaES(\vv_i;\vtheta, \vphi)^\top \frac{\partial\vf_\veta(\vv_i)}{\partial \veta} \nonumber \\
    & + \frac{\partial \vepsilon_i^\top \nabla_\vv \vf_\veta(\vv_i) \vepsilon_i}{\partial \veta} - \frac{1}{2} \frac{\partial ||\vf_\veta (\vv_i)||_2^2}{\partial \veta} \big],
\end{align}

where $\vv_{1:M}$ is a mini-batch from $p_D(\vv)$ and $\vepsilon_{1:M}$ is a mini-batch from $p(\vepsilon)$. We refer to this method as VaGES-Fisher.

\vspace{-.1cm}
\subsection{Experiments}
\begin{figure}[t]
\begin{center}
\includegraphics[width=0.85\columnwidth]{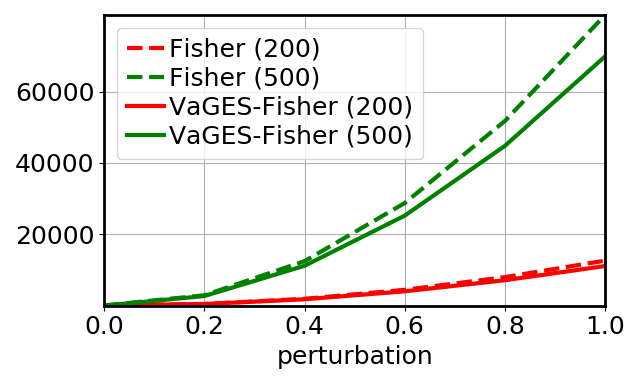}
\vspace{-.3cm}
\caption{The Fisher divergence estimated by VaGES-Fisher v.s. the accurate Fisher divergence (Fisher).}
\label{fig:evaluate_grbm}
\vspace{-.5cm}
\end{center}
\end{figure}

We validate the effectiveness of VaGES-Fisher in GRBMs. We initialize a GRBM as $p_D(\vv)$, perturb its weight with increasing noise and estimate the Fisher divergence between the initial GRBM and the perturbed one by VaGES-Fisher. The dimensions of $\vv$ and $\vh$ are the same and we experiment on dimensions of 200 and 500. See Appendix~\ref{sec:add_fisher} for more experimental details. The result is shown in Fig.~\ref{fig:evaluate_grbm}. We compare the Fisher divergence estimated by VaGES-Fisher with the accurate Fisher divergence. Under both dimensions, our estimated Fisher divergence is close to the accurate one.

\vspace{-.1cm}
\section{Related Work}
\label{sec:related_work}
\textbf{Learning EBLVMs with MLE.} A popular class of learning methods is the maximum likelihood estimate (MLE). Such methods have to deal with both the model posterior and the partition function. Some methods impose structural assumptions on the model. Among them, contrastive divergence (CD)~\cite{hinton2002training} and its variants~\cite{tieleman2008training,yixuan2020unbiased} estimate the gradient of the partition function with respect to the model parameters using Gibbs sampling~\cite{geman1984stochastic} when the posterior can be sampled efficiently. Furthermore, when the model is fully visible, scalable methods~\cite{du2018learning,du2019implicit,nijkamp2019anatomy,nijkamp2019learning,grathwohl2019your} are applicable by estimating such gradient with Langevin dynamics. In deep models with multiple layers of latent variables such as DBNs~\cite{hinton2006fast,lee2009convolutional} and DBMs~\cite{salakhutdinov2009deep}, the greedy layer-wise learning algorithm~\cite{hinton2006fast,bengio2007greedy} is applicable by exploring their hierarchical structures. Some recent methods~\cite{kuleshov2017neural,li2019relieve} attempt to learn general EBLVMs by variational inference. However, such methods suffer from high variance~\cite{kuleshov2017neural} or high bias~\cite{li2019relieve} of the variational bounds for the partition function in high-dimensional space.

\textbf{Learning EBLVMs without partition functions.} As mentioned in Sec.~\ref{sec:intro}, the score matching (SM) is a popular class of learning methods, which eliminate the partition function by considering the score function, and the recent bi-level score matching (BiSM) is a scalable extension of SM to learn general EBLVMs. Prior to BiSM, extensions of SM~\cite{swersky2011autoencoders,verteslearning} impose structural assumptions on the model, e.g., a tractable posterior~\cite{swersky2011autoencoders} or a jointly exponential family model~\cite{verteslearning}. Recently, \citet{song2019generative,song2020improved} use a score network to directly model the score function of the data distribution. Incorporating latent variables to such a model is currently an open problem and we leave it as future work. The noise contrastive estimation (NCE)~\cite{gutmann2010noise} is also a learning method which eliminates the partition function and variational noise contrastive estimation (VNCE)~\cite{pmlr-v89-rhodes19a} is an extension of NCE to general EBLVMs based on variational methods. However, such methods have to manually design a noise distribution, which in principle should be close to the data distribution and is challenging in high-dimensional space. There are alternative approaches to handling the partition function, e.g., Doubly Dual Embedding~\cite{dai2019kernel}, Adversarial Dynamics Embedding~\cite{dai2019exponential} and Fenchel Mini-Max Learning~\cite{tao2019fenchel}, and extending theses methods for learning EBLVMs would be interesting future work.

\textbf{Evaluating EBLVMs.} The log-likelihood is commonly used to compare fully visible models on the same dataset and the annealed importance sampling~\cite{neal2001annealed,salakhutdinov2008learning} is often used to estimate the partition function in the log-likelihood. For a general EBLVM, a lower bound of the log-likelihood~\cite{kingma2013auto,burda2015importance} can be estimated by introducing a variational posterior together with estimating the partition function. The log-likelihood is a relative value and the corresponding absolute value (i.e., the KL divergence) is generally intractable since the entropy of the data distribution is unknown.


\section{Conclusion}
We propose new variational estimates of the score function and its gradient with respect to the model parameters in general energy-based latent variable models (EBLVMs). We rewrite the score function and its gradient as combinations of expectation and covariance terms over the model posterior. We approximate the model posterior with a variational posterior and analyze its bias. With samples drawn from the variational posterior, the expectation terms are estimated by Monte Carlo and the covariance terms are estimated by sample covariance matrix. We apply our estimates to kernelized Stein discrepancy (KSD), score matching (SM)-based methods and exact Fisher divergence. In particular, these estimates applied to SM-based methods are more time and memory efficient compared with the complex gradient unrolling in the strongest baseline bi-level score matching (BiSM) and meanwhile can scale up to natural images in deep EBLVMs. Besides, we also present interpolation results of annealed Langevin dynamics trajectories in the latent space in deep EBLVMs.

\section*{Software and Data}
Codes in \url{https://github.com/baofff/VaGES}.


\section*{Acknowledgements}
This work was supported by NSFC Projects (Nos.  61620106010, 62061136001, 61621136008, U1811461, 62076145), Beijing NSF Project (No. JQ19016), Beijing Academy of Artificial Intelligence (BAAI), Tsinghua-Huawei Joint Research Program, a grant from Tsinghua Institute for Guo Qiang, Tiangong Institute for Intelligent Computing, and the NVIDIA NVAIL Program with GPU/DGX Acceleration. C. Li was supported by the fellowship of China postdoctoral Science Foundation (2020M680572), and the fellowship of China national postdoctoral program for innovative talents (BX20190172) and Shuimu Tsinghua Scholar.




\bibliography{example_paper}
\bibliographystyle{icml2021}

\newpage
\appendix
\onecolumn

\section{Proof of the Decomposition of the Gradient of the Score Function}
\label{sec:proof_eqn_grad_score}
\begin{proof}
Firstly we have
\vspace{-.1cm}
\begin{align}
\label{eqn:app_zero_mean}
\E_{p_\vtheta(\vh|\vv)} \left[ \nabla_\vv \log p_\vtheta(\vh|\vv) \right] = & \E_{p_\vtheta(\vh|\vv)} \left[\frac{\nabla_\vv p_\vtheta(\vh|\vv)}{p_\vtheta(\vh|\vv)}\right] = \int p_\vtheta(\vh|\vv) \frac{\nabla_\vv p_\vtheta(\vh|\vv)}{p_\vtheta(\vh|\vv)} \mathrm{d}\vh \nonumber \\
= & \int \nabla_\vv p_\vtheta(\vh|\vv) \mathrm{d}\vh = \nabla_\vv \int p_\vtheta(\vh|\vv) \mathrm{d}\vh = \nabla_\vv 1 = \vzero,
\end{align}
\vspace{-.1cm}
and similarly we have $\label{eqn:app_zero_mean_theta}
\E_{p_\vtheta(\vh|\vv)} \left[ \nabla_\vtheta \log p_\vtheta(\vh|\vv) \right] = \vzero$. Thereby, we have
\vspace{-.1cm}
\begin{align*}
    \nabla_\vv \log \tilde{p}_\vtheta(\vv) = \nabla_\vv \log \tilde{p}_\vtheta(\vv) + \E_{p_\vtheta(\vh|\vv)} \left[ \nabla_\vv \log p_\vtheta(\vh|\vv) \right] = \E_{p_\vtheta(\vh|\vv)} \left[ \nabla_{\vv} \log \tilde{p}_\vtheta(\vv, \vh) \right],
\end{align*}
\vspace{-.1cm}
and similarly we have $\nabla_\vtheta \log \tilde{p}_\vtheta(\vv) = \E_{p_\vtheta(\vh|\vv)} \left[ \nabla_{\vtheta} \log \tilde{p}_\vtheta(\vv, \vh) \right]$.

Taking derivatives to Eqn.~(\ref{eqn:app_zero_mean}) w.r.t. $\vtheta$, we have
\vspace{-.1cm}
\begin{align}
\label{eqn:app_sd}
    \E_{p_\vtheta(\vh|\vv)} \left[ \frac{\partial \nabla_\vv \log p_\vtheta(\vh|\vv)}{\partial \vtheta} \right] + \E_{p_\vtheta(\vh|\vv)} \left[ \nabla_\vv \log p_\vtheta(\vh|\vv) \frac{\partial \log p_\vtheta(\vh|\vv)}{\partial \vtheta} \right] = \vzero.
\end{align}
\vspace{-.1cm}
The second term in the left side of Eqn.~(\ref{eqn:app_sd}) can be written as
\vspace{-.1cm}
\begin{align*}
    & \E_{p_\vtheta(\vh|\vv)} \left[ \nabla_\vv \log p_\vtheta(\vh|\vv) \frac{\partial \log p_\vtheta(\vh|\vv)}{\partial \vtheta} \right] \\
    = & \E_{p_\vtheta(\vh|\vv)} \left[ \nabla_\vv \log p_\vtheta(\vv, \vh) \frac{\partial \log p_\vtheta(\vh|\vv)}{\partial \vtheta} \right] - \E_{p_\vtheta(\vh|\vv)} \left[ \nabla_\vv \log p_\vtheta(\vv) \frac{\partial \log p_\vtheta(\vh|\vv)}{\partial \vtheta} \right] \\
    = & \E_{p_\vtheta(\vh|\vv)} \left[ \nabla_\vv \log p_\vtheta(\vv, \vh) \frac{\partial \log p_\vtheta(\vh|\vv)}{\partial \vtheta} \right] \\
    = & \E_{p_\vtheta(\vh|\vv)} \left[ \nabla_\vv \log p_\vtheta(\vv, \vh) \frac{\partial \log \tilde{p}_\vtheta(\vv, \vh)}{\partial \vtheta} \right] - \E_{p_\vtheta(\vh|\vv)} \left[ \nabla_\vv \log p_\vtheta(\vv, \vh) \frac{\partial \log \tilde{p}_\vtheta(\vv)}{\partial \vtheta} \right] \\
    = & \E_{p_\vtheta(\vh|\vv)} \left[ \nabla_\vv \log p_\vtheta(\vv, \vh) \frac{\partial \log \tilde{p}_\vtheta(\vv, \vh)}{\partial \vtheta} \right] - \E_{p_\vtheta(\vh|\vv)} \left[ \nabla_\vv \log p_\vtheta(\vv, \vh) \right] \frac{\partial \log \tilde{p}_\vtheta(\vv)}{\partial \vtheta} \\
    = & \E_{p_\vtheta(\vh|\vv)} \left[ \nabla_\vv \log p_\vtheta(\vv, \vh) \frac{\partial \log \tilde{p}_\vtheta(\vv, \vh)}{\partial \vtheta} \right] - \E_{p_\vtheta(\vh|\vv)} \left[ \nabla_\vv \log p_\vtheta(\vv, \vh) \right] \E_{p_\vtheta(\vh|\vv)} \left[ \frac{\partial \log \tilde{p}_\vtheta(\vv, \vh)}{\partial \vtheta} \right] \\
    = & \Cov_{p_\vtheta(\vh|\vv)}(\nabla_\vv \log p_\vtheta(\vv, \vh), \nabla_\vtheta \log \tilde{p}_\vtheta(\vv, \vh)) = \Cov_{p_\vtheta(\vh|\vv)}(\nabla_\vv \log \tilde{p}_\vtheta(\vv, \vh), \nabla_\vtheta \log \tilde{p}_\vtheta(\vv, \vh)).
\end{align*}

Thereby, we have
\begin{align*}
    & \frac{\partial \nabla_\vv \log p_\vtheta(\vv)}{\partial \vtheta} = \frac{\partial \nabla_\vv \log \tilde{p}_\vtheta(\vv)}{\partial \vtheta} \\
    = & \frac{\partial \nabla_\vv \log \tilde{p}_\vtheta(\vv)}{\partial \vtheta} + \E_{p_\vtheta(\vh|\vv)} \left[ \frac{\partial \nabla_\vv \log p_\vtheta(\vh|\vv)}{\partial \vtheta} \right] + \E_{p_\vtheta(\vh|\vv)} \left[ \nabla_\vv \log p_\vtheta(\vh|\vv) \frac{\partial \log p_\vtheta(\vh|\vv)}{\partial \vtheta} \right] \\
    = & \E_{p_\vtheta(\vh|\vv)} \left[ \frac{\partial \nabla_\vv \log \tilde{p}_\vtheta(\vv, \vh)}{\partial \vtheta} \right] + \E_{p_\vtheta(\vh|\vv)} \left[ \nabla_\vv \log p_\vtheta(\vh|\vv) \frac{\partial \log p_\vtheta(\vh|\vv)}{\partial \vtheta} \right] \\
    = & \E_{p_\vtheta(\vh|\vv)} \left[ \frac{\partial \nabla_\vv \log \tilde{p}_\vtheta(\vv, \vh)}{\partial \vtheta} \right] + \Cov_{p_\vtheta(\vh|\vv)}(\nabla_\vv \log \tilde{p}_\vtheta(\vv, \vh), \nabla_\vtheta \log \tilde{p}_\vtheta(\vv, \vh)). \qedhere
\end{align*}
\end{proof}

\section{The Tractability of the Commonly Used Divergences between Posteriors}
\label{sec:tract}
The tractability of the KL divergence or the Fisher divergence between the variational posterior and the true posterior in EBLVMs has been shown by~\citet{bao2020bilevel} and we restate the results for completeness. Besides, we also analyze the tractability of the reverse KL divergence, the total variation distance~\cite{pollard2005inequalities}, the maximum mean discrepancy~\cite{li2017mmd} and the Wasserstein distance~\cite{arjovsky2017wasserstein} between the two posteriors in EBLVMs.

\textbf{The KL divergence is tractable.} The gradient of the KL divergence between $q_\vphi(\vh|\vv)$ and $p_\vtheta(\vh|\vv)$ w.r.t. $\vphi$ is
\begin{align*}
    \nabla_\vphi \gD_{KL}(q_\vphi(\vh|\vv)||p_\vtheta(\vh|\vv)) = & \nabla_\vphi \E_{q_\vphi(\vh|\vv)} \log \frac{q_\vphi(\vh|\vv)}{p_\vtheta(\vh|\vv)} = \nabla_\vphi \E_{q_\vphi(\vh|\vv)} \log \frac{q_\vphi(\vh|\vv) p_\vtheta(\vv) \gZ(\vtheta)}{\tilde{p}_\vtheta(\vv, \vh)} \\
    = & \nabla_\vphi \E_{q_\vphi(\vh|\vv)} \log \frac{q_\vphi(\vh|\vv)}{\tilde{p}_\vtheta(\vv, \vh)} + \cancel{\nabla_\vphi \log p_\vtheta(\vv) \gZ(\vtheta)} = \nabla_\vphi \E_{q_\vphi(\vh|\vv)} \log \frac{q_\vphi(\vh|\vv)}{\tilde{p}_\vtheta(\vv, \vh)}.
\end{align*}
The last term doesn't depend on the partition function or the marginal distribution of an EBLVM and thereby is tractable.

\textbf{The Fisher divergence is tractable.} The Fisher divergence between $q_\vphi(\vh|\vv)$ and $p_\vtheta(\vh|\vv)$ is
\begin{align*}
    \gD_F(q_\vphi(\vh|\vv)||p_\vtheta(\vh|\vv)) = & \frac{1}{2} \E_{q_\vphi(\vh|\vv)} || \nabla_\vh \log q_\vphi(\vh|\vv) - \nabla_\vh \log p_\vtheta(\vh|\vv) ||_2^2 \\
    = & \frac{1}{2} \E_{q_\vphi(\vh|\vv)} || \nabla_\vh \log q_\vphi(\vh|\vv) - \nabla_\vh \log \frac{\tilde{p}_\vtheta(\vv, \vh)}{p_\vtheta(\vv) \gZ(\vtheta)} ||_2^2 \\
    = & \frac{1}{2} \E_{q_\vphi(\vh|\vv)} || \nabla_\vh \log q_\vphi(\vh|\vv) - \nabla_\vh \log \tilde{p}_\vtheta(\vv, \vh) - \cancel{\nabla_\vh \log p_\vtheta(\vv) \gZ(\vtheta)} ||_2^2 \\
    = & \frac{1}{2} \E_{q_\vphi(\vh|\vv)} || \nabla_\vh \log q_\vphi(\vh|\vv) - \nabla_\vh \log \tilde{p}_\vtheta(\vv, \vh)||_2^2.
\end{align*}
Again, the last term doesn't depend on the partition function or the marginal distribution of an EBLVM and thereby is tractable. Furthermore, its gradient w.r.t. $\vphi$ is tractable.

\textbf{The reverse KL divergence is generally intractable.} The gradient of the reverse KL divergence between $q_\vphi(\vh|\vv)$ and $p_\vtheta(\vh|\vv)$ w.r.t. $\vphi$ is
$\nabla_\vphi \gD_{RKL}(q_\vphi(\vh|\vv)||p_\vtheta(\vh|\vv)) = \nabla_\vphi \E_{p_\vtheta(\vh|\vv)} \log \frac{p_\vtheta(\vh|\vv)}{q_\vphi(\vh|\vv)} = - \E_{p_\vtheta(\vh|\vv)} \nabla_\vphi \log q_\vphi(\vh|\vv)$.
Since $p_\vtheta(\vh|\vv)$ is generally intractable, the reverse KL divergence is generally intractable.

\textbf{The total variation distance is generally intractable.} The gradient of total variation distance~\cite{pollard2005inequalities} between $q_\vphi(\vh|\vv)$ and $p_\vtheta(\vh|\vv)$ w.r.t. $\vphi$ is
$\nabla_\vphi V(q_\vphi(\vh|\vv), p_\vphi(\vh|\vv)) = \nabla_\vphi \frac{1}{2} \E_{q_\vphi(\vh|\vv)} |1 - \frac{p_\vtheta(\vh|\vv)}{q_\vphi(\vh|\vv)}|$.
Since $p_\vtheta(\vh|\vv)$ is generally intractable, the total variation distance is generally intractable.

\textbf{Maximum mean discrepancy is generally intractable.} Given a kernel $k$, the gradient of the square of maximum mean discrepancy~\cite{li2017mmd} between $q_\vphi(\vh|\vv)$ and $p_\vtheta(\vh|\vv)$ w.r.t. $\vphi$ is
\begin{align*}
    \nabla_\vphi M_k (q_\vphi(\vh|\vv), p_\vtheta(\vh|\vv)) & = \nabla_\vphi \big( \E_{\vh, \vh' \sim q_\vphi(\vh|\vv)} k(\vh, \vh') + \E_{\vh, \vh' \sim p_\vtheta(\vh|\vv)} k(\vh, \vh') - 2 \E_{\vh \sim q_\vphi(\vh|\vv), \vh' \sim p_\vtheta(\vh'|\vv)} k(\vh, \vh') \big) \\
    & = \nabla_\vphi \E_{\vh, \vh' \sim q_\vphi(\vh|\vv)} k(\vh, \vh') - 2 \E_{\vh' \sim p_\vtheta(\vh'|\vv)} \nabla_\vphi \E_{\vh \sim q_\vphi(\vh|\vv)} k(\vh, \vh').
\end{align*}
Since $p_\vtheta(\vh|\vv)$ is generally intractable, the maximum mean discrepancy is generally intractable.

\textbf{The Wasserstein distance is generally intractable.} The Wasserstein  distance~\cite{arjovsky2017wasserstein} between $q_\vphi(\vh|\vv)$ and $p_\vtheta(\vh|\vv)$ is
$W(q_\vphi(\vh|\vv), p_\vtheta(\vh|\vv)) = \frac{1}{K} \sup\limits_{f: ||f||_{Lip} \leq K} \E_{q_\vphi(\vh|\vv)} f(\vh) - \E_{p_\vtheta(\vh|\vv)} f(\vh)$.
Generally $\{f: ||f||_{Lip} \leq K\}$ is approximated by a neural network $f_\veta$  with weight clipping and the Wasserstein distance is optimized by a bi-level optimization~\cite{arjovsky2017wasserstein}. The lower level problem requires samples from the generally intractable posterior $p_\vtheta(\vh|\vv)$. Thereby, the Wasserstein distance is generally intractable.

\vspace{-.1cm}
\section{Proof of Theorem 1 and Theorem 2 }
\label{sec:proof_bd_kl}




\begin{customlemma}{1}
\label{thm:kl_bound}
Suppose $P, Q$ are two probability measures on $\Omega$, and $\vf: \Omega \rightarrow \sR^m$, then we have $||\E_P \vf - \E_Q \vf||_2 \leq ||\vf||_\infty \sqrt{2 \gD_{KL}(Q||P)}$,
where $||\vf||_\infty \triangleq \sup\limits_{\omega\in \Omega} ||\vf(\omega)||_2$.
\vspace{-.4cm}
\begin{proof}
Let $S = (P+Q) / 2$, then $P, Q$ are absolutely continuous w.r.t. $S$, and we have
\begin{align*}
||\E_P \vf - \E_Q \vf||_2 = & ||\int \vf \mathrm{d} P - \int \vf \mathrm{d} Q ||_2 = ||\int \vf \frac{\mathrm{d} P}{\mathrm{d} S} \mathrm{d} S - \int \vf \frac{\mathrm{d} Q}{\mathrm{d} S} \mathrm{d} S ||_2 = ||\int \vf (\frac{\mathrm{d} P}{\mathrm{d} S} - \frac{\mathrm{d} Q}{\mathrm{d} S}) \mathrm{d} S ||_2 \\
\leq & \int ||\vf||_2\ |\frac{\mathrm{d} P}{\mathrm{d} S} - \frac{\mathrm{d} Q}{\mathrm{d} S}| \mathrm{d} S \leq ||\vf||_\infty \int |\frac{\mathrm{d} P}{\mathrm{d} S} - \frac{\mathrm{d} Q}{\mathrm{d} S}| \mathrm{d} S.
\end{align*}
According to Pinsker's inequality~\citep{tsybakov2008introduction}, we have $\int |\frac{\mathrm{d} P}{\mathrm{d} S} - \frac{\mathrm{d} Q}{\mathrm{d} S}| \mathrm{d} S \leq \sqrt{2 \gD_{KL}(Q||P)}$. Thereby, $||\E_P \vf - \E_Q \vf||_2 \leq ||\vf||_\infty \sqrt{2 \gD_{KL}(Q||P)}$.
\end{proof}
\end{customlemma}

\begin{customthm}{1}
\label{thm:app_bd_vaes_kl}
(VaES, KL divergence) Suppose $\nabla_{\vv} \log \tilde{p}_\vtheta(\vv, \vh)$ is bounded w.r.t. $\vv, \vh$ and $\vtheta$, then the bias of $\mathrm{VaES}(\vv; \vtheta, \vphi)$ can be bounded by the square root of the KL divergence between $q_\vphi(\vh|\vv)$ and $p_\vtheta(\vh|\vv)$ up to multiplying a constant.
\vspace{-.2cm}
\begin{proof}
According to Lemma~\ref{thm:kl_bound}, we have
\begin{align*}
    ||\E_{q_\vphi(\vh|\vv)} \left[ \nabla_\vv \log \tilde{p}_\vtheta (\vv, \vh) \right] - \E_{p_\vtheta(\vh|\vv)} \left[ \nabla_\vv \log \tilde{p}_\vtheta (\vv, \vh) \right]||_2
    \leq \sup_{\vh} ||\nabla_\vv \log \tilde{p}_\vtheta(\vv, \vh)||_2 \sqrt{2 \gD_{KL} (q_\vphi(\vh|\vv) || p_\vtheta(\vh|\vv)) }.
\end{align*}
By the boundedness of $\nabla_{\vv} \log \tilde{p}_\vtheta(\vv, \vh)$, $\exists A < \infty, \forall \vv, \forall \vh, \forall \vtheta, ||\nabla_{\vv} \log \tilde{p}_\vtheta(\vv, \vh)||_2 \leq A$. Let $C = \sqrt{2} A$, then
\begin{align*}
    & ||\E_{q_\vphi(\vh|\vv)} \left[ \nabla_\vv \log \tilde{p}_\vtheta (\vv, \vh) \right] - \nabla_\vv \log p_\vtheta (\vv)||_2 = ||\E_{q_\vphi(\vh|\vv)} \left[ \nabla_\vv \log \tilde{p}_\vtheta (\vv, \vh) \right] - \E_{p_\vtheta(\vh|\vv)} \left[ \nabla_\vv \log \tilde{p}_\vtheta (\vv, \vh) \right]||_2 \\
    \leq & A \sqrt{2 \gD_{KL} (q_\vphi(\vh|\vv) || p_\vtheta(\vh|\vv)) } = C \sqrt{\gD_{KL} (q_\vphi(\vh|\vv) || p_\vtheta(\vh|\vv)) }. \qedhere
\end{align*}
\end{proof}
\end{customthm}

\begin{customdef}{1}
Suppose $A$ is a matrix, we define $||A||_2 \triangleq \sqrt{\sum\limits_{i,j} A_{i,j}^2}$.
\end{customdef}

\begin{customlemma}{2}
\label{thm:matrix_norm}
Suppose $\va$, $\vb$ are two vectors, then $||\va \vb^\top||_2 = ||\va||_2 ||\vb||_2$.
\begin{proof}
$||\va \vb^\top||_2 = \sqrt{\sum\limits_{i,j}a_i^2 b_j^2} = ||\va||_2 ||\vb||_2$.
\end{proof}
\end{customlemma}

\begin{customthm}{2}
\label{thm:app_bd_vages_kl}
(VaGES, KL divergence) Suppose $\nabla_{\vv} \log \tilde{p}_\vtheta(\vv, \vh)$, $\nabla_{\vtheta} \log \tilde{p}_\vtheta(\vv, \vh)$ and $\frac{\partial \nabla_\vv \log \tilde{p}_\vtheta (\vv, \vh)}{\partial \vtheta}$ are bounded w.r.t. $\vv, \vh$ and $\vtheta$, then the bias of $\mathrm{VaGES}(\vv; \vtheta, \vphi)$ can be bounded by the square root of the KL divergence between $q_\vphi(\vh|\vv)$ and $p_\vtheta(\vh|\vv)$ up to multiplying a constant.
\vspace{-.2cm}
\begin{proof}
According to Thm.~\ref{thm:app_bd_vaes_kl}, $\exists C_1 < \infty$, s.t.
\begin{align*}
    ||\E_{q_\vphi(\vh|\vv)} \left[ \nabla_\vv \log \tilde{p}_\vtheta (\vv, \vh) \right] - \nabla_\vv \log \tilde{p}_\vtheta (\vv)||_2
    \leq C_1 \sqrt{\gD_{KL} (q_\vphi(\vh|\vv) || p_\vtheta(\vh|\vv)) }.
\end{align*}

Similarly, $\exists C_2 < \infty$, s.t.
\begin{align*}
    ||\E_{q_\vphi(\vh|\vv)} \left[ \nabla_\vtheta \log \tilde{p}_\vtheta (\vv, \vh) \right] - \nabla_\vtheta \log \tilde{p}_\vtheta (\vv)||_2
    \leq C_2 \sqrt{\gD_{KL} (q_\vphi(\vh|\vv) || p_\vtheta(\vh|\vv)) },
\end{align*}
and $\exists C_3 < \infty$, s.t.
\begin{align*}
    ||\E_{q_\vphi(\vh|\vv)} \left[ \frac{\partial \nabla_\vv \log \tilde{p}_\vtheta (\vv, \vh)}{\partial \vtheta} \right] - \E_{p_\vtheta(\vh|\vv)} \left[ \frac{\partial \nabla_\vv \log \tilde{p}_\vtheta (\vv, \vh)}{\partial \vtheta} \right]||_2 \leq C_3 \sqrt{\gD_{KL} (q_\vphi(\vh|\vv) || p_\vtheta(\vh|\vv)) }.
\end{align*}

By the boundedness of $\nabla_\vv \log \tilde{p}_\vtheta(\vv, \vh)$ and $\nabla_\vtheta \log \tilde{p}_\vtheta(\vv, \vh)$, $\nabla_\vv \log \tilde{p}_\vtheta(\vv, \vh) \frac{\partial \log \tilde{p}_\vtheta(\vv, \vh)}{\partial \vtheta}$ is also bounded. Thereby, $\exists C_4 < \infty$, s.t.
\begin{align*}
    & ||\E_{q_\vphi(\vh|\vv)} \left[ \nabla_\vv \log \tilde{p}_\vtheta(\vv, \vh) \frac{\partial \log \tilde{p}_\vtheta(\vv, \vh)}{\partial \vtheta} \right] - \E_{p_\vtheta(\vh|\vv)} \left[ \nabla_\vv \log \tilde{p}_\vtheta(\vv, \vh) \frac{\partial \log \tilde{p}_\vtheta(\vv, \vh)}{\partial \vtheta} \right]||_2 \\
    \leq & C_4 \sqrt{\gD_{KL} (q_\vphi(\vh|\vv) || p_\vtheta(\vh|\vv)) }.
\end{align*}

By the boundedness of $\nabla_{\vv} \log \tilde{p}_\vtheta(\vv, \vh)$ and $\nabla_{\vtheta} \log \tilde{p}_\vtheta(\vv, \vh)$, we can assume $C < \infty$ is a constant that bounds $||\nabla_{\vv} \log \tilde{p}_\vtheta(\vv, \vh)||_2$ and $||\nabla_{\vtheta} \log \tilde{p}_\vtheta(\vv, \vh)||_2$. Then, by the triangle inequality and Lemma.~\ref{thm:matrix_norm}, we have
\begin{align*}
    &||\E_{q_\vphi(\vh|\vv)} \left[ \nabla_\vv \log \tilde{p}_\vtheta(\vv, \vh) \right] \E_{q_\vphi(\vh|\vv)} \left[ \frac{\partial \log \tilde{p}_\vtheta(\vv, \vh)}{\partial \vtheta} \right] -\E_{p_\vtheta(\vh|\vv)} \left[ \nabla_\vv \log \tilde{p}_\vtheta(\vv, \vh) \right] \E_{p_\vtheta(\vh|\vv)} \left[ \frac{\partial \log \tilde{p}_\vtheta(\vv, \vh)}{\partial \vtheta} \right]||_2 \\
    \leq & ||\E_{q_\vphi(\vh|\vv)} \left[ \nabla_\vv \log \tilde{p}_\vtheta(\vv, \vh) \right] \left(\E_{q_\vphi(\vh|\vv)} \left[ \frac{\partial \log \tilde{p}_\vtheta(\vv, \vh)}{\partial \vtheta}\right] - \E_{p_\vtheta(\vh|\vv)} \left[ \frac{\partial \log \tilde{p}_\vtheta(\vv, \vh)}{\partial \vtheta}\right] \right)||_2 \\
    &+ || \left(\E_{q_\vphi(\vh|\vv)} \left[ \nabla_\vv \log \tilde{p}_\vtheta(\vv, \vh) \right] - \E_{p_\vtheta(\vh|\vv)} \left[ \nabla_\vv \log \tilde{p}_\vtheta(\vv, \vh)\right] \right) \E_{p_\vtheta(\vh|\vv)} \left[ \frac{\partial \log \tilde{p}_\vtheta(\vv, \vh)}{\partial \vtheta} \right]||_2 \\
    = & ||\E_{q_\vphi(\vh|\vv)} \left[ \nabla_\vv \log \tilde{p}_\vtheta(\vv, \vh) \right]||_2\  ||\E_{q_\vphi(\vh|\vv)} \left[ \frac{\partial \log \tilde{p}_\vtheta(\vv, \vh)}{\partial \vtheta} \right] - \E_{p_\vtheta(\vh|\vv)} \left[ \frac{\partial \log \tilde{p}_\vtheta(\vv, \vh)}{\partial \vtheta} \right]||_2 \\
    &+ || \E_{q_\vphi(\vh|\vv)} \left[ \nabla_\vv \log \tilde{p}_\vtheta(\vv, \vh) \right] - \E_{p_\vtheta(\vh|\vv)} \left[ \nabla_\vv \log \tilde{p}_\vtheta(\vv, \vh) \right] ||_2 \  ||\E_{p_\vtheta(\vh|\vv)} \left[ \frac{\partial \log \tilde{p}_\vtheta(\vv, \vh)}{\partial \vtheta} \right]||_2 \\
    \leq & (C C_2 + C C_1) \sqrt{\gD_{KL} (q_\vphi(\vh|\vv) || p_\vtheta(\vh|\vv)) }.
\end{align*}

Thereby,
\begin{align*}
    & ||\mathrm{Cov}_{q_\vphi(\vh|\vv)}(\nabla_\vv \log \tilde{p}_\vtheta(\vv, \vh), \nabla_\vtheta \log \tilde{p}_\vtheta(\vv, \vh)) - \mathrm{Cov}_{p_\vtheta(\vh|\vv)}(\nabla_\vv \log \tilde{p}_\vtheta(\vv, \vh), \nabla_\vtheta \log \tilde{p}_\vtheta(\vv, \vh))||_2 \\
    \leq & ||\E_{q_\vphi(\vh|\vv)} \left[ \nabla_\vv \log \tilde{p}_\vtheta(\vv, \vh) \frac{\partial \log \tilde{p}_\vtheta(\vv, \vh)}{\partial \vtheta} \right] - \E_{p_\vtheta(\vh|\vv)} \left[ \nabla_\vv \log \tilde{p}_\vtheta(\vv, \vh) \frac{\partial \log \tilde{p}_\vtheta(\vv, \vh)}{\partial \vtheta} \right]||_2 \\
    & + ||\E_{q_\vphi(\vh|\vv)} \left[ \nabla_\vv \log \tilde{p}_\vtheta(\vv, \vh) \right] \E_{q_\vphi(\vh|\vv)} \left[ \frac{\partial \log \tilde{p}_\vtheta(\vv, \vh)}{\partial \vtheta} \right]-\E_{p_\vtheta(\vh|\vv)} \left[ \nabla_\vv \log \tilde{p}_\vtheta(\vv, \vh) \right] \E_{p_\vtheta(\vh|\vv)} \left[ \frac{\partial \log \tilde{p}_\vtheta(\vv, \vh)}{\partial \vtheta} \right]||_2 \\
    \leq & (C_4 + C C_2 + C C_1) \sqrt{\gD_{KL} (q_\vphi(\vh|\vv) || p_\vtheta(\vh|\vv)) }.
\end{align*}

As a result,
\begin{align*}
& ||\mathrm{Cov}_{q_\vphi(\vh|\vv)}(\nabla_\vv \log \tilde{p}_\vtheta(\vv, \vh), \nabla_\vtheta \log \tilde{p}_\vtheta(\vv, \vh))+ \E_{q_\vphi(\vh|\vv)} \left[\frac{\partial \nabla_\vv \log \tilde{p}_\vtheta(\vv, \vh)}{\partial \vtheta}\right] - \frac{\partial \nabla_{\vv}\log p_\vtheta(\vv)}{\partial \vtheta}||_2\\
= & ||\mathrm{Cov}_{q_\vphi(\vh|\vv)}(\nabla_\vv \log \tilde{p}_\vtheta(\vv, \vh), \nabla_\vtheta \log \tilde{p}_\vtheta(\vv, \vh)) + \E_{q_\vphi(\vh|\vv)} \left[\frac{\partial \nabla_\vv \log \tilde{p}_\vtheta(\vv, \vh)}{\partial \vtheta}\right] \\
& - \mathrm{Cov}_{p_\vtheta(\vh|\vv)}(\nabla_\vv \log \tilde{p}_\vtheta(\vv, \vh), \nabla_\vtheta \log \tilde{p}_\vtheta(\vv, \vh)) - \E_{p_\vtheta(\vh|\vv)} \left[\frac{\partial \nabla_\vv \log \tilde{p}_\vtheta(\vv, \vh)}{\partial \vtheta}\right]||_2 \\
\leq & ||\mathrm{Cov}_{q_\vphi(\vh|\vv)}(\nabla_\vv \log \tilde{p}_\vtheta(\vv, \vh), \nabla_\vtheta \log \tilde{p}_\vtheta(\vv, \vh)) - \mathrm{Cov}_{p_\vtheta(\vh|\vv)}(\nabla_\vv \log \tilde{p}_\vtheta(\vv, \vh), \nabla_\vtheta \log \tilde{p}_\vtheta(\vv, \vh))||_2 \\
& + ||\E_{q_\vphi(\vh|\vv)} \left[\frac{\partial \nabla_\vv \log \tilde{p}_\vtheta(\vv, \vh)}{\partial \vtheta}\right] - \E_{p_\vtheta(\vh|\vv)} \left[\frac{\partial \nabla_\vv \log \tilde{p}_\vtheta(\vv, \vh)}{\partial \vtheta}\right] ||_2\\
\leq & (C_4 + C C_2 + C C_1 + C_3) \sqrt{\gD_{KL} (q_\vphi(\vh|\vv) || p_\vtheta(\vh|\vv)) }. \qedhere
\end{align*}

\end{proof}
\end{customthm}

\section{Proof of Theorem 3 and Theorem 4}
\label{sec:proof_bd_fisher}

\begin{customdef}{2}
Suppose $p$ is a probability density on $\sR^n$ and $\vg: \sR^n \rightarrow \sR^n$, we define $\gS_p \vg(\vx) \triangleq \nabla_\vx \log p(\vx)^\top \vg(\vx) + \Tr (\nabla_\vx \vg(\vx))$.
\end{customdef}

\begin{customlemma}{3}
\label{thm:stein_id}
\citep{liu2016stein} Suppose $p$ is a probability density on $\sR^n$ and $\vg: \sR^n \rightarrow \sR^n$ is a function satisfying $\lim\limits_{||\vx|| \rightarrow \infty} p(\vx) \vg(\vx) = \vzero$, then $\E_{p(\vx)} \left[ \gS_p \vg(\vx) \right] = 0$.
\vspace{-.2cm}
\begin{proof}
\begin{align*}
\vzero = \int \nabla_\vx (p(\vx) \vg(\vx)) \mathrm{d} \vx = \int p(\vx) \nabla_\vx \vg(\vx) + p(\vx) \vg(\vx) \nabla_\vx \log p(\vx)^\top \mathrm{d} \vx = \E_{p(\vx)} \left[ \nabla_\vx \vg (\vx) + \vg(\vx) \nabla_\vx \log p(\vx)^\top \right].
\end{align*}
Thereby,
\begin{align*}
&0 = \Tr (\E_{p(\vx)} \left[ \nabla_\vx \vg (\vx) + \vg(\vx) \nabla_\vx \log p(\vx)^\top \right]) = \E_{p(\vx)} \left[ \Tr(\nabla_\vx \vg (\vx)) +  \nabla_\vx \log p(\vx)^\top \vg(\vx) \right] = \E_{p(\vx)} \left[ \gS_p \vg(\vx) \right]. \qedhere
\end{align*}
\end{proof}
\end{customlemma}

\begin{customlemma}{4}
\label{thm:fisher_bound_a}
Suppose $p, q$ are probability densities on $\sR^n$ and $\vg: \sR^n \rightarrow \sR^n$ satisfies $\lim\limits_{||\vx|| \rightarrow \infty} q(\vx) \vg(\vx) = \vzero$, we have 
\begin{align*}
    |\E_q \gS_p \vg| \leq \sqrt{\E_{q(\vx)} ||\vg(\vx)||^2} \sqrt{D_F(q||p)}
\end{align*}
\begin{proof}
By Lemma~\ref{thm:stein_id}, we have $\E_q \gS_q \vg = 0$. Thereby,
\begin{align*}
    & |\E_q \gS_p \vg| = |\E_q \gS_p \vg - \E_q \gS_q \vg| = |\E_{q(\vx)} \vg (\vx)^\top (\nabla_\vx \log p(\vx) - \nabla_\vx \log q(\vx))| \\
    \leq & \E_{q(\vx)} ||\vg(\vx)||\ ||\nabla_\vx \log p(\vx) - \nabla_\vx \log q(\vx)|| \leq \sqrt{\E_{q(\vx)} ||\vg(\vx)||^2 \E_{q(\vx)}||\nabla_\vx \log p(\vx) - \nabla_\vx \log q(\vx)||^2} \\
    = & \sqrt{\E_{q(\vx)} ||\vg(\vx)||^2} \sqrt{D_F(q||p)}. \qedhere
\end{align*}
\end{proof}
\end{customlemma}

\begin{customdef}{3}
{\rm\citep{ley2013stein}} Suppose $p$ is a probability density defined on $\sR^n$ and $f: \sR^n \rightarrow \sR$ is a function, we define $\vg_f^p$ as a solution of the Stein equation $\gS_p \vg = f - \E_p f$.
\end{customdef}

\begin{remark}
The solution of the Stein equation exists. For example, let $h = f - \E_p f$, then
\begin{align*}
    g_1(\vx) = \frac{1}{p(\vx)} \int_{-\infty}^{x_1}p(t, x_2,\cdots,x_n)h(t,x_2,\cdots,x_n)\mathrm{d}t,\quad g_2(\vx) = \cdots = g_n(\vx) = 0
\end{align*}
is a solution.
\end{remark}

\begin{customdef}{4}
Suppose $p, q$ are probability densities defined on $\sR^n$ and $\vf: \sR^n \rightarrow \sR^m$ is a function, we say $\vf$ satisfies the Stein regular condition w.r.t. $p, q$ iff $\forall i \in \sZ \cap [1, m], \lim\limits_{||\vx|| \rightarrow \infty} q(\vx) \vg_{f_i}^p (\vx) =0 $.
\end{customdef}

\begin{customdef}{5}
{\rm\citep{ley2013stein}} Suppose $p, q$ are probability densities defined on $\sR^n$ and $\vf: \sR^n \rightarrow \sR^m$ is a function satisfying the Stein regular condition w.r.t. $p, q$, we define $\kappa_\vf^{p,q} \triangleq \sqrt{\E_{q(\vx)}\sum\limits_{i=1}^m ||\vg_{f_i}^p (\vx)||_2^2}$, referred to as the Stein factor of $\vf$ w.r.t. $p, q$.
\end{customdef}

\begin{customlemma}{5}
\label{thm:fisher_bound}
Suppose $p, q$ are probability densities defined on $\sR^n$ and $\vf: \sR^n \rightarrow \sR^m$ is a function satisfying the Stein regular condition w.r.t. $p, q$, then we have $|| \E_q \vf - \E_p \vf||_2 \leq \kappa_\vf^{p,q} \sqrt{\gD_F(q||p)}$.
\vspace{-.2cm}
\begin{proof}
By Lemma~\ref{thm:fisher_bound_a}, we have
$|\E_q f_i - \E_p f_i| = |\E_q (f_i - \E_p f_i)| = |\E_q \gS_p \vg_{f_i}^p| \leq \sqrt{\E_{q(\vx)} ||\vg_{f_i}^p (\vx)||_2^2} \sqrt{D_F(q||p)}$.

Thereby, $|| \E_q \vf - \E_p \vf|| = \sqrt{\sum\limits_{i=1}^n |\E_q f_i - \E_p f_i|^2 } \leq \sqrt{\sum\limits_{i=1}^n \E_{q(\vx)} ||\vg_{f_i}^p (\vx)||_2^2 \ D_F(q||p) } = \kappa_\vf^{p,q} \sqrt{D_F(q||p)}$.
\end{proof}
\end{customlemma}

\begin{customthm}{3}
\label{thm:app_bd_vaes_fisher}
(continuous $\vh$, VaES) Suppose {\rm(1)} $\forall (\vv, \vtheta, \vphi)$, $\nabla_\vv \log \tilde{p}_\vtheta(\vv, \vh)$ as a function of $\vh$ satisfies the Stein regular condition w.r.t. $p_\vtheta(\vh|\vv)$ and $q_\vphi(\vh|\vv)$ and {\rm(2)} the Stein factor of $\nabla_\vv \log \tilde{p}_\vtheta(\vv, \vh)$ as a function of $\vh$ w.r.t. $p_\vtheta(\vh|\vv), q_\vphi(\vh|\vv)$ is bounded w.r.t. $\vv, \vtheta$ and $\vphi$, then the bias of $\mathrm{VaES}(\vv; \vtheta, \vphi)$ can be bounded by the square root of the Fisher divergence between $q_\vphi(\vh|\vv)$ and $p_\vtheta(\vh|\vv)$ up to multiplying a constant.
\begin{proof}
\vspace{-.2cm}
It can be directly derived from Lemma~\ref{thm:fisher_bound}.
\end{proof}
\end{customthm}

\begin{customthm}{4}
(continuous $\vh$, VaGES) Suppose {\rm(1)} $\forall (\vv, \vtheta, \vphi)$, $\nabla_\vv \log \tilde{p}_\vtheta(\vv, \vh)$, $\nabla_\vtheta \log \tilde{p}_\vtheta(\vv, \vh)$, $\nabla_\vv \log \tilde{p}_\vtheta(\vv, \vh) \frac{\partial \log \tilde{p}_\vtheta(\vv, \vh)}{\partial \vtheta}$ and $\frac{\partial \nabla_\vv \log \tilde{p}_\vtheta(\vv, \vh)}{\partial \vtheta}$ as functions of $\vh$ satisfy the Stein regular condition w.r.t. $p_\vtheta(\vh|\vv)$ and $q_\vphi(\vh|\vv)$ and {\rm(2)} the Stein factors of $\nabla_\vv \log \tilde{p}_\vtheta(\vv, \vh)$, $\nabla_\vtheta \log \tilde{p}_\vtheta(\vv, \vh)$, $\nabla_\vv \log \tilde{p}_\vtheta(\vv, \vh) \frac{\partial \log \tilde{p}_\vtheta(\vv, \vh)}{\partial \vtheta}$ and $\frac{\partial \nabla_\vv \log \tilde{p}_\vtheta(\vv, \vh)}{\partial \vtheta}$ as functions of $\vh$ w.r.t. $p_\vtheta(\vh|\vv), q_\vphi(\vh|\vv)$ are bounded w.r.t. $\vv, \vtheta$ and $\vphi$, {\rm(3)} $\nabla_{\vv} \log \tilde{p}_\vtheta(\vv, \vh)$ and $\nabla_{\vtheta} \log \tilde{p}_\vtheta(\vv, \vh)$ are bounded w.r.t. $\vv, \vh$ and $\vtheta$, then the bias of $\mathrm{VaGES}(\vv; \vtheta, \vphi)$ can be bounded by the square root of the Fisher divergence between $q_\vphi(\vh|\vv)$ and $p_\vtheta(\vh|\vv)$ up to multiplying a constant.
\vspace{-.2cm}
\begin{proof}
According to Lemma~\ref{thm:fisher_bound}, $\exists C_1 < \infty$, s.t.
\begin{align*}
    ||\E_{q_\vphi(\vh|\vv)} \left[ \nabla_\vv \log \tilde{p}_\vtheta (\vv, \vh) \right] - \nabla_\vv \log \tilde{p}_\vtheta (\vv)||_2
    \leq C_1 \sqrt{\gD_{F} (q_\vphi(\vh|\vv) || p_\vtheta(\vh|\vv)) },
\end{align*}

$\exists C_2 < \infty$, s.t.
\begin{align*}
    ||\E_{q_\vphi(\vh|\vv)} \left[ \nabla_\vtheta \log \tilde{p}_\vtheta (\vv, \vh) \right] - \nabla_\vtheta \log \tilde{p}_\vtheta (\vv)||_2
    \leq C_2 \sqrt{\gD_{F} (q_\vphi(\vh|\vv) || p_\vtheta(\vh|\vv)) },
\end{align*}
$\exists C_3 < \infty$, s.t.
\begin{align*}
    & ||\E_{q_\vphi(\vh|\vv)} \left[ \frac{\partial \nabla_\vv \log \tilde{p}_\vtheta (\vv, \vh)}{\partial \vtheta} \right] - \E_{p_\vtheta(\vh|\vv)} \left[ \frac{\partial \nabla_\vv \log \tilde{p}_\vtheta (\vv, \vh)}{\partial \vtheta} \right]||_2 
    \leq  C_3 \sqrt{\gD_{F} (q_\vphi(\vh|\vv) || p_\vtheta(\vh|\vv)) },
\end{align*}

and $\exists C_4 < \infty$, s.t.
\begin{align*}
    & ||\E_{q_\vphi(\vh|\vv)} \left[ \nabla_\vv \log \tilde{p}_\vtheta(\vv, \vh) \frac{\partial \log \tilde{p}_\vtheta(\vv, \vh)}{\partial \vtheta} \right] - \E_{p_\vtheta(\vh|\vv)} \left[ \nabla_\vv \log \tilde{p}_\vtheta(\vv, \vh) \frac{\partial \log \tilde{p}_\vtheta(\vv, \vh)}{\partial \vtheta} \right]||_2 \\
    \leq & C_4 \sqrt{\gD_{F} (q_\vphi(\vh|\vv) || p_\vtheta(\vh|\vv)) }.
\end{align*}

By the boundedness of $\nabla_{\vv} \log \tilde{p}_\vtheta(\vv, \vh)$ and $\nabla_{\vtheta} \log \tilde{p}_\vtheta(\vv, \vh)$, we can assume $C < \infty$ is a constant that bounds $||\nabla_{\vv} \log \tilde{p}_\vtheta(\vv, \vh)||_2$ and $||\nabla_{\vtheta} \log \tilde{p}_\vtheta(\vv, \vh)||_2$. After establishing the above bounds w.r.t. the Fisher divergence, the rest proof is exactly the same as Theorem~\ref{thm:app_bd_vages_kl}. For completeness
, we restate the proof as follows. By the triangle inequality and Lemma.~\ref{thm:matrix_norm}, we have
\begin{align*}
    &||\E_{q_\vphi(\vh|\vv)} \left[ \nabla_\vv \log \tilde{p}_\vtheta(\vv, \vh) \right] \E_{q_\vphi(\vh|\vv)} \left[ \frac{\partial \log \tilde{p}_\vtheta(\vv, \vh)}{\partial \vtheta} \right]-\E_{p_\vtheta(\vh|\vv)} \left[ \nabla_\vv \log \tilde{p}_\vtheta(\vv, \vh) \right] \E_{p_\vtheta(\vh|\vv)} \left[ \frac{\partial \log \tilde{p}_\vtheta(\vv, \vh)}{\partial \vtheta} \right]||_2 \\
    \leq & ||\E_{q_\vphi(\vh|\vv)} \left[ \nabla_\vv \log \tilde{p}_\vtheta(\vv, \vh) \right] \left(\E_{q_\vphi(\vh|\vv)} \left[ \frac{\partial \log \tilde{p}_\vtheta(\vv, \vh)}{\partial \vtheta}\right] - \E_{p_\vtheta(\vh|\vv)} \left[ \frac{\partial \log \tilde{p}_\vtheta(\vv, \vh)}{\partial \vtheta}\right] \right)||_2 \\
    &+ || \left(\E_{q_\vphi(\vh|\vv)} \left[ \nabla_\vv \log \tilde{p}_\vtheta(\vv, \vh) \right] - \E_{p_\vtheta(\vh|\vv)} \left[ \nabla_\vv \log \tilde{p}_\vtheta(\vv, \vh)\right] \right) \E_{p_\vtheta(\vh|\vv)} \left[ \frac{\partial \log \tilde{p}_\vtheta(\vv, \vh)}{\partial \vtheta} \right]||_2 \\
    = & ||\E_{q_\vphi(\vh|\vv)} \left[ \nabla_\vv \log \tilde{p}_\vtheta(\vv, \vh) \right]||_2\  ||\E_{q_\vphi(\vh|\vv)} \left[ \frac{\partial \log \tilde{p}_\vtheta(\vv, \vh)}{\partial \vtheta} \right] - \E_{p_\vtheta(\vh|\vv)} \left[ \frac{\partial \log \tilde{p}_\vtheta(\vv, \vh)}{\partial \vtheta} \right]||_2 \\
    &+ || \E_{q_\vphi(\vh|\vv)} \left[ \nabla_\vv \log \tilde{p}_\vtheta(\vv, \vh) \right] - \E_{p_\vtheta(\vh|\vv)} \left[ \nabla_\vv \log \tilde{p}_\vtheta(\vv, \vh) \right] ||_2 \  ||\E_{p_\vtheta(\vh|\vv)} \left[ \frac{\partial \log \tilde{p}_\vtheta(\vv, \vh)}{\partial \vtheta} \right]||_2 \\
    \leq & (C C_2 + C C_1) \sqrt{\gD_{F} (q_\vphi(\vh|\vv) || p_\vtheta(\vh|\vv)) }.
\end{align*}

Thereby,
\begin{align*}
    & ||\mathrm{Cov}_{q_\vphi(\vh|\vv)}(\nabla_\vv \log \tilde{p}_\vtheta(\vv, \vh), \nabla_\vtheta \log \tilde{p}_\vtheta(\vv, \vh)) - \mathrm{Cov}_{p_\vtheta(\vh|\vv)}(\nabla_\vv \log \tilde{p}_\vtheta(\vv, \vh), \nabla_\vtheta \log \tilde{p}_\vtheta(\vv, \vh))||_2 \\
    \leq & ||\E_{q_\vphi(\vh|\vv)} \left[ \nabla_\vv \log \tilde{p}_\vtheta(\vv, \vh) \frac{\partial \log \tilde{p}_\vtheta(\vv, \vh)}{\partial \vtheta} \right] - \E_{p_\vtheta(\vh|\vv)} \left[ \nabla_\vv \log \tilde{p}_\vtheta(\vv, \vh) \frac{\partial \log \tilde{p}_\vtheta(\vv, \vh)}{\partial \vtheta} \right]||_2 \\
    & + ||\E_{q_\vphi(\vh|\vv)} \left[ \nabla_\vv \log \tilde{p}_\vtheta(\vv, \vh) \right] \E_{q_\vphi(\vh|\vv)} \left[ \frac{\partial \log \tilde{p}_\vtheta(\vv, \vh)}{\partial \vtheta} \right]-\E_{p_\vtheta(\vh|\vv)} \left[ \nabla_\vv \log \tilde{p}_\vtheta(\vv, \vh) \right] \E_{p_\vtheta(\vh|\vv)} \left[ \frac{\partial \log \tilde{p}_\vtheta(\vv, \vh)}{\partial \vtheta} \right]||_2 \\
    \leq & (C_4 + C C_2 + C C_1) \sqrt{\gD_{F} (q_\vphi(\vh|\vv) || p_\vtheta(\vh|\vv)) }.
\end{align*}

As a result,
\begin{align*}
& ||\mathrm{Cov}_{q_\vphi(\vh|\vv)}(\nabla_\vv \log \tilde{p}_\vtheta(\vv, \vh), \nabla_\vtheta \log \tilde{p}_\vtheta(\vv, \vh))+ \E_{q_\vphi(\vh|\vv)} \left[\frac{\partial \nabla_\vv \log \tilde{p}_\vtheta(\vv, \vh)}{\partial \vtheta}\right] - \frac{\partial \nabla_{\vv}\log p_\vtheta(\vv)}{\partial \vtheta}||_2\\
= & ||\mathrm{Cov}_{q_\vphi(\vh|\vv)}(\nabla_\vv \log \tilde{p}_\vtheta(\vv, \vh), \nabla_\vtheta \log \tilde{p}_\vtheta(\vv, \vh)) + \E_{q_\vphi(\vh|\vv)} \left[\frac{\partial \nabla_\vv \log \tilde{p}_\vtheta(\vv, \vh)}{\partial \vtheta}\right] \\
& - \mathrm{Cov}_{p_\vtheta(\vh|\vv)}(\nabla_\vv \log \tilde{p}_\vtheta(\vv, \vh), \nabla_\vtheta \log \tilde{p}_\vtheta(\vv, \vh)) - \E_{p_\vtheta(\vh|\vv)} \left[\frac{\partial \nabla_\vv \log \tilde{p}_\vtheta(\vv, \vh)}{\partial \vtheta}\right]||_2 \\
\leq & ||\mathrm{Cov}_{q_\vphi(\vh|\vv)}(\nabla_\vv \log \tilde{p}_\vtheta(\vv, \vh), \nabla_\vtheta \log \tilde{p}_\vtheta(\vv, \vh)) - \mathrm{Cov}_{p_\vtheta(\vh|\vv)}(\nabla_\vv \log \tilde{p}_\vtheta(\vv, \vh), \nabla_\vtheta \log \tilde{p}_\vtheta(\vv, \vh))||_2 \\
& + ||\E_{q_\vphi(\vh|\vv)} \left[\frac{\partial \nabla_\vv \log \tilde{p}_\vtheta(\vv, \vh)}{\partial \vtheta}\right] - \E_{p_\vtheta(\vh|\vv)} \left[\frac{\partial \nabla_\vv \log \tilde{p}_\vtheta(\vv, \vh)}{\partial \vtheta}\right] ||_2\\
\leq & (C_4 + C C_2 + C C_1 + C_3) \sqrt{\gD_{F} (q_\vphi(\vh|\vv) || p_\vtheta(\vh|\vv)) }.  \qedhere
\end{align*}
\end{proof}
\end{customthm}

\vspace{-.2cm}
\section{Consistency between $\gD_F^m$ and $\gD_F$}
\label{sec:consist}
\begin{customthm}{5}
Suppose $\lim\limits_{||\vv||\rightarrow \infty} p_D(\vv) (\nabla_\vv \log p_\vtheta (\vv) - \nabla_\vv \log p_D (\vv)) = \vzero$ and $\E_{p(\vepsilon)} \left[\vepsilon \vepsilon^\top\right] = \bm{I}$, then $\gD_F^m (p_D || p_\vtheta) = \gD_F (p_D || p_\vtheta)$, where $\gD_F (p_D || p_\vtheta) = \frac{1}{2} \E_{p_D(\vv)} || \nabla_{\vv} \log p_\vtheta(\vv)\! -\! \nabla_{\vv}  \log p_D(\vv) ||_2^2$ is the Fisher divergence between $p_D$ and $p_\vtheta$.
\vspace{-.4cm}
\begin{proof}
By the assumption $\E_{p(\vepsilon)} \left[\vepsilon \vepsilon^\top\right] = \bm{I}$, we have $\E_{p(\vepsilon)} \left[ \vepsilon^\top \nabla_\vv \vf (\vv) \vepsilon \right] = \Tr (\nabla_\vv \vf (\vv))$. Thereby,
$\gD_F^m (p_D || p_\vtheta) = \max\limits_{\vf \in \gF} \E_{p_D(\vv)} \left[ \nabla_\vv \log p_\vtheta(\vv)^\top \vf(\vv) + \Tr ( \nabla_\vv \vf(\vv) ) - \frac{1}{2}||\vf(\vv)||_2^2 \right]$.

Suppose $\vf \in \gF$, i.e., $\vf$ is a function from $\sR^d$ to $\sR^d$ and $\lim\limits_{||\vv|| \rightarrow \infty} p_D(\vv) \vf(\vv) = \vzero$, by the Stein's identity, we have $\E_{p_D(\vv)} \left[ \nabla_\vv \log p_D(\vv)^\top \vf(\vv) + \Tr (\nabla_\vv \vf(\vv)) \right] = \vzero$. Thereby, we have
\begin{align*}
    & \E_{p_D(\vv)} \left[ \nabla_\vv \log p_\vtheta(\vv)^\top \vf(\vv) + \Tr (\nabla_\vv \vf(\vv)) - \frac{1}{2}||\vf(\vv)||_2^2 \right] \\
    = & \E_{p_D(\vv)} \left[ \nabla_\vv \log p_\vtheta(\vv)^\top \vf(\vv) + \Tr (\nabla_\vv \vf(\vv)) - \frac{1}{2}||\vf(\vv)||_2^2 \right] - \E_{p_D(\vv)} \left[ \nabla_\vv \log p_D(\vv)^\top \vf(\vv) + \Tr (\nabla_\vv \vf(\vv)) \right] \\
    = & \E_{p_D(\vv)} \left[ (\nabla_\vv \log p_\vtheta(\vv) - \nabla_\vv \log p_D (\vv) )^\top \vf(\vv) - \frac{1}{2}||\vf(\vv)||_2^2 \right] \\
    \leq & \frac{1}{2} \E_{p_D(\vv)} \left[ ||\nabla_\vv \log p_\vtheta(\vv) - \nabla_\vv \log p_D (\vv) )||_2^2 \right] = \gD_F(p_D || p_\vtheta).
\end{align*}
The equality is achieved when $\vf(\vv) = \nabla_\vv \log p_\vtheta(\vv) - \nabla_\vv \log p_D (\vv)$, which is a function in $\gF$ by assumption. As a result, $\gD_F^m (p_D || p_\vtheta) = \gD_F (p_D || p_\vtheta)$.
\end{proof}
\end{customthm}

\newpage

\section{Additional Experimental Details}
\label{sec:setting}

\subsection{Learning EBLVMs with KSD}
\label{sec:add_setting_ksd}
\textbf{Additional setting.} We generate 60,000 samples for training and 10,000 samples for testing on checkerboard. The dimension of $\vh$ is 4. We use the Adam optimizer and the learning rate is $10^{-3}$. We train 100,000 iterations and the batch size is 100. The log-likelihood is estimated by annealed importance sampling~\cite{salakhutdinov2008quantitative}, where we use 2,000 samples and 2,000 middle states to estimate the log-partition function. We run 1,000 steps Gibbs sampling to sample from GRBMs. The variational parameter $\vphi$ is updated for $K=5$ times on each minibatch. 

\subsection{Learning EBLVMs with Score Matching}

\subsubsection{Comparison in GRBMs}
\label{sec:setting_cmp_grbm}
Following BiSM~\cite{bao2020bilevel}, we split 1,400 images for training, 300 images for validation and 265 images for testing on Frey face\footnote{http://www.cs.nyu.edu/\textasciitilde roweis/data.html}; we generate 60,000 samples for training and 10,000 samples for testing on checkerboard; the dimension of $\vh$ is 400 on Frey face and 4 on checkerboard; we use the Adam optimizers with learning rates $2\times 10^{-4}$ for training on Frey face and $10^{-3}$ for training on checkerboard; we train 20,000 iterations on Frey face and 100,000 iterations on checkerboard; the batch size is 100 for both datasets; we select the best model trained on Frey face according to the validation log-likelihood. The log-likelihood is estimated by annealed importance sampling~\cite{salakhutdinov2008quantitative}, where we use 2,000 samples and 2,000 middle states to estimate the log-partition function. We run 1,000 steps Gibbs sampling to sample from GRBMs. The time comparison on Frey face is conducted on 1 GeForce GTX 1080 Ti GPU with 2,000 training iterations. The number of samples from $q_\vphi(\vh|\vv)$ is $L=2$ and the variational parameter $\vphi$ is updated for $K=5$ times on each minibatch by default.

\subsubsection{Learning Deep EBLVMs}
\label{sec:setting_learn_eblvm}

\begin{wrapfigure}{R}{0.38\textwidth}
\begin{minipage}{\linewidth}
\vspace{-1cm}
\begin{figure}[H]
\centering
    \includegraphics[width=\linewidth]{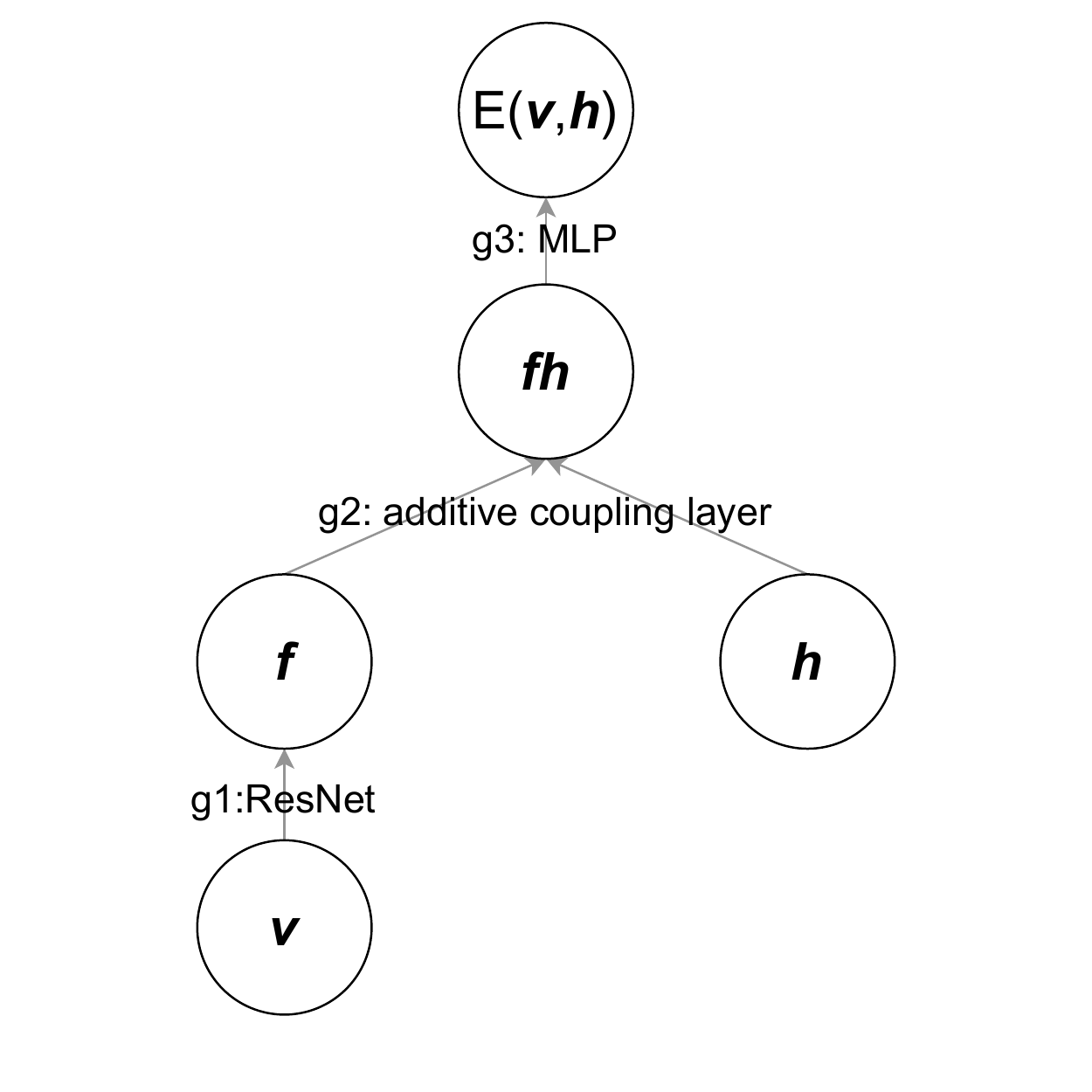}
    \vspace{-1cm}
    \caption{The structure of the deep EBLVM.}
    \label{fig:net}
\end{figure}
\end{minipage}
\end{wrapfigure}
\textbf{Additional setting.} Following BiSM~\cite{bao2020bilevel}, we split 60,000 samples for training on MNIST, 50,000 samples for training on CIFAR10 and 182,637 samples for training on CelebA; the dimension of $\vh$ is 50; we use the Adam optimizers with learning rates $10^{-4}$ for training on MNIST and $5\times 10^{-5}$ for training on CIFAR10 and CelebA; we train 100,000 iterations on MNIST and 300,000 iterations on CIFAR10 and CelebA; the batch size is 100 for all datasets. $g_1$ consists of a $6k$-layer ResNet, where $k=2,3,3$ for MNIST, CIFAR10 and CelebA respectively and $g_3$ is an MLP containing one fully connected layer. The structure of the deep EBLVM is shown in Fig.~\ref{fig:net}. The number of samples from $q_\vphi(\vh|\vv)$ is $L=2$ and the variational parameter $\vphi$ is updated for $K=5$ times on each minibatch. Following a similar protocol with ~\citet{song2019generative,li2019annealed,bao2020bilevel}, we save one checkpoint every 5000 iterations and select the best CIFAR10 and CelebA models according to the FID score on 1000 samples. The FID score reported in Section~\ref{sec:deep_eblvm} is estimated on 50,000 samples using the official code\footnote{https://github.com/bioinf-jku/TTUR}.

\textbf{Hyperparamter selection.} Since we compare with BiSM~\cite{bao2020bilevel}, we use the same hyperparameters as BiSM when they can be shared (e.g., the model types and structures, the divergence to learn $q_\vphi(\vh|\vv)$, the dimensions of $\vh$, the batch size, the optimizers and corresponding learning rates). As for $L$ (the number of samples from $q_\vphi(\vh|\vv)$, we find it enough to set it to 2 (the minimal number of samples required in a sample covariance matrix) in our considered models. As for $K$, we set it to 5, so that it will ensure the convergence of training $q_\vphi (\vh|\vv)$ and meanwhile have an acceptable computation cost. As for the step size and the standard deviation of the noise in Langevin dynamics, we grid search the optimal one, as shown in Tab.~\ref{tab:grid_search}. We find that the optimal step size is approximately proportional to the dimension of $\vh$, perhaps because Langevin dynamics converges to its stationary distribution slower when $\vh$ has a higher dimension. The standard deviation can work in range $[10^{-4}, 10^{-2}]$.

\begin{table}[H]
\centering
\caption{Grid search of the step size and the standard deviation of the noise in Langevin dynamics under different dimensions of $\vh$. We use the CIFAR10 dataset. The result is represented by the FID score on 1000 samples.}
\label{tab:grid_search}
\subfloat[$\mathrm{dimension}(\vh) = 20$]{\begin{tabular}{cccc}
\toprule
      & $10^{-4}$ & $10^{-3}$ & $10^{-2}$\\
\midrule
    $1\times 10^{-3}$  & diverge & diverge & diverge \\
    $2\times 10^{-3}$  & 56.42 & 54.87 & 54.46 \\
    $\mathbf{4\times 10^{-3}}$  & 55.90 & 58.07 & 54.55\\
    $6\times 10^{-3}$  & 56.48 & 55.09 & 57.03\\
    $10\times 10^{-3}$  & diverge & diverge & diverge\\
\bottomrule
\end{tabular}}\quad
\subfloat[$\mathrm{dimension}(\vh) = 50$]{\begin{tabular}{cccc}
\toprule
      & $10^{-4}$ & $10^{-3}$ & $10^{-2}$\\
\midrule
    $2.5\times 10^{-3}$ & diverge & diverge & diverge \\
    $5\times 10^{-3}$ & 56.74 & 57.55 & 58.19 \\
    $\mathbf{10\times 10^{-3}}$ & 55.74 & 54.71 & 58.01 \\
    $15\times 10^{-3}$ & 58.03 & 60.52 & 56.18 \\
    $25\times 10^{-3}$  & diverge & diverge & diverge \\
\bottomrule
\end{tabular}}\\
\subfloat[$\mathrm{dimension}(\vh) = 100$]{\begin{tabular}{cccc}
\toprule
      & $10^{-4}$ & $10^{-3}$ & $10^{-2}$\\
\midrule
    $10 \times 10^{-3}$ & diverge & diverge & diverge \\
    $\mathbf{20\times 10^{-3}}$ & 56.57 & 60.04 & 56.31 \\
    $30\times 10^{-3}$ & diverge & diverge & diverge \\
\bottomrule
\end{tabular}}
\end{table}

\textbf{Sampling.} Since we compare with BiSM~\cite{bao2020bilevel}, we use the same sampling algorithm as BiSM. For deep EBLVMs, we first randomly select a training data point and inference its approximate posterior mean; we then sample from $p(\vv|\vh)$ with $\vh$ equal to the approximate posterior mean using the annealed Langevin dynamics technique~\cite{li2019annealed}. The temperature range is $[1, 100]$ and the step size is $0.02$ in annealed Langevin dynamics.

\textbf{Devices and training time.} The time and memory consumption of VaGES with different batch sizes (BS) is displayed in Tab.~\ref{tab:tm}. We also include that of BiSM. The time and memory consumption in deep EBLVMs of VaGES and BiSM is consistent with GRBMs (see Fig.~\ref{fig:freyface_cmp} in the full paper). Besides, training an EBLVM takes about 2.8 times as long as training an EBM (the one trained by MDSM in Tab.~{\ref{tab:fid} (a)} in the full paper). The additional cost is reasonable, since (i) the EBLVM improves the expressive power (see Tab.~{\ref{tab:fid} (a)} in the full paper) based on a similar model structure and a comparable amount of parameters (244MB for the EBLVM and 238MB for the EBM), and (ii) the EBLVM enables manipulation in the latent space (see Fig.~\ref{fig:trajectory} in the full paper). 
\begin{table}[H]
\vspace{-0.6cm}
    \centering
    \caption{Training time of 2k iterations/memory consumption on GeForce RTX 2080 Ti in deep EBLVMs. $L$=2,$K$=5.}
    \label{tab:tm}
    \begin{tabular}{cccccc}
    \toprule
        Dataset & Setting & VaGES & BiSM ($N$=0) & BiSM ($N$=2) & BiSM ($N$=5) \\
    \midrule
        \multirow{2}{*}{CIFAR10} & 1GPU BS=64 & 24m/5.8GB & 19m/6.8GB & 25m/7.8GB & 34m/9.5GB \\
        & 2GPUs BS=100 & 35m/8.2GB & 28m/11.1GB & 43m/13.2GB & 66m/16.4GB \\
    \midrule
        \multirow{2}{*}{CelebA} & 1GPU BS=16 & 26m/8.0GB & 21m/7.9GB & 26m/9.2GB & 35m/10.2GB \\
        & 6GPUs BS=100 & 90m/52.4GB & 67m/51.9GB & 100m/58.7GB & 124m/60.1GB\\
    \bottomrule
    \end{tabular}
\vspace{-.4cm}
\end{table}

\subsection{Evaluating EBLVMs with Exact Fisher Divergence}
\label{sec:add_fisher}
\textbf{Additional setting.} The GRBM is initialized as a standard Gaussian distribution by letting $\vb = \vzero, \vc = \vzero, W=\vzero, \sigma=1$, so we can get accurate samples from it. We get 20,000 samples from the initial GRBM, and split 16,000 samples for training, 2,000 samples for validation and 2,000 samples for testing. We use the Adam optimizer and the learning rate is $2\times 10^{-4}$. We train 20,000 iterations and the batch size is 100. The number of samples from $q_\vphi(\vh|\vv)$ is $L=1$ and the variational parameter $\vphi$ is updated for $K=5$ times on each minibatch. $q_\vphi(\vh|\vv)$ is a Bernoulli distribution parametermized by a fully connected layer with the sigmoid activation and we use the Gumbel-Softmax trick~\citep{jang2016categorical} for reparameterization of $q_\vphi(\vh|\vv)$ with 0.1 as the temperature. $\gD$ is the KL divergence to learn $q_\vphi(\vh|\vv)$. $\vf_\veta$ is a multilayer perceptron (MLP) with 2 hidden layers and each layer has the same width.

\subsection{Numerical Validation of Theorems}
\label{sec:numer}
In the two posteriors $p_\vtheta(\vh|\vv)$ and $q_\vphi(\vh|\vv)$, we fix $\vv$, $\vtheta$ and only vary $\vphi$ to plot the relationship between the biases and the divergences. As for $\vv$, we randomly select a sample from the Frey face training dataset and fix $\vv$ as the sample. As for $\vtheta$, we randomly initialize it with the uniform noise and don't change it anymore. As for $\vphi$, we initialize it such that the variational posterior $q_\vphi(\vh|\vv)$ is equal to the true posterior $p_\vtheta(\vh|\vv)$. After initialization, we perturb $\vphi$ with an increasing Gaussian noise level and record the corresponding biases and divergences.

The dimension of $\vh$ is 400. As for the GRBM, $q_\vphi(\vh|\vv)$ is a Bernoulli distribution parametermized by a fully connected layer with the sigmoid activation. As for the Gaussian model, $q_\vphi(\vh|\vv)$ is a Gaussian distribution parametermized by a fully connected layer.

\section{Additional Results}
\subsection{Learning EBLVMs with KSD}
\label{sec:add_ksd}
\begin{figure}[H]
\begin{center}
\subfloat[Data density]{\includegraphics[width=0.19\columnwidth]{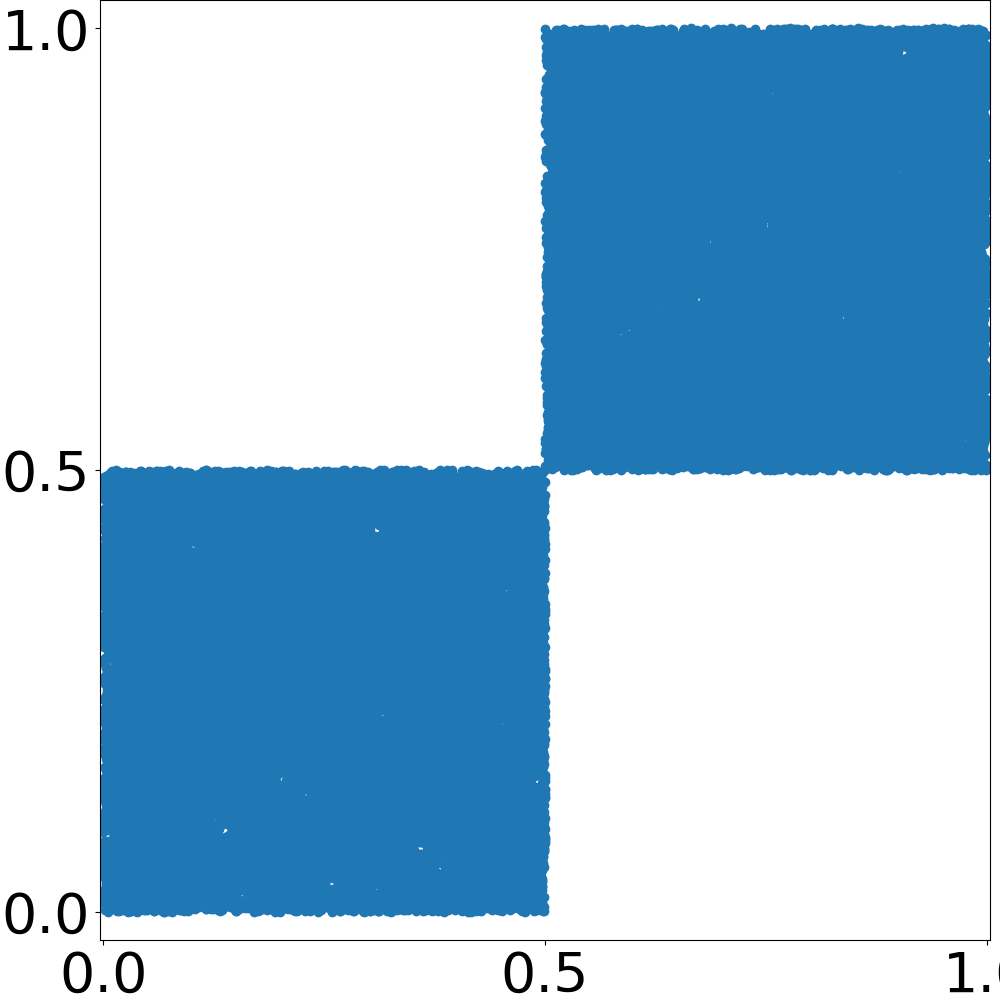}}\quad
\subfloat[VaGES-KSD ($L$=2)]{\includegraphics[width=0.19\columnwidth]{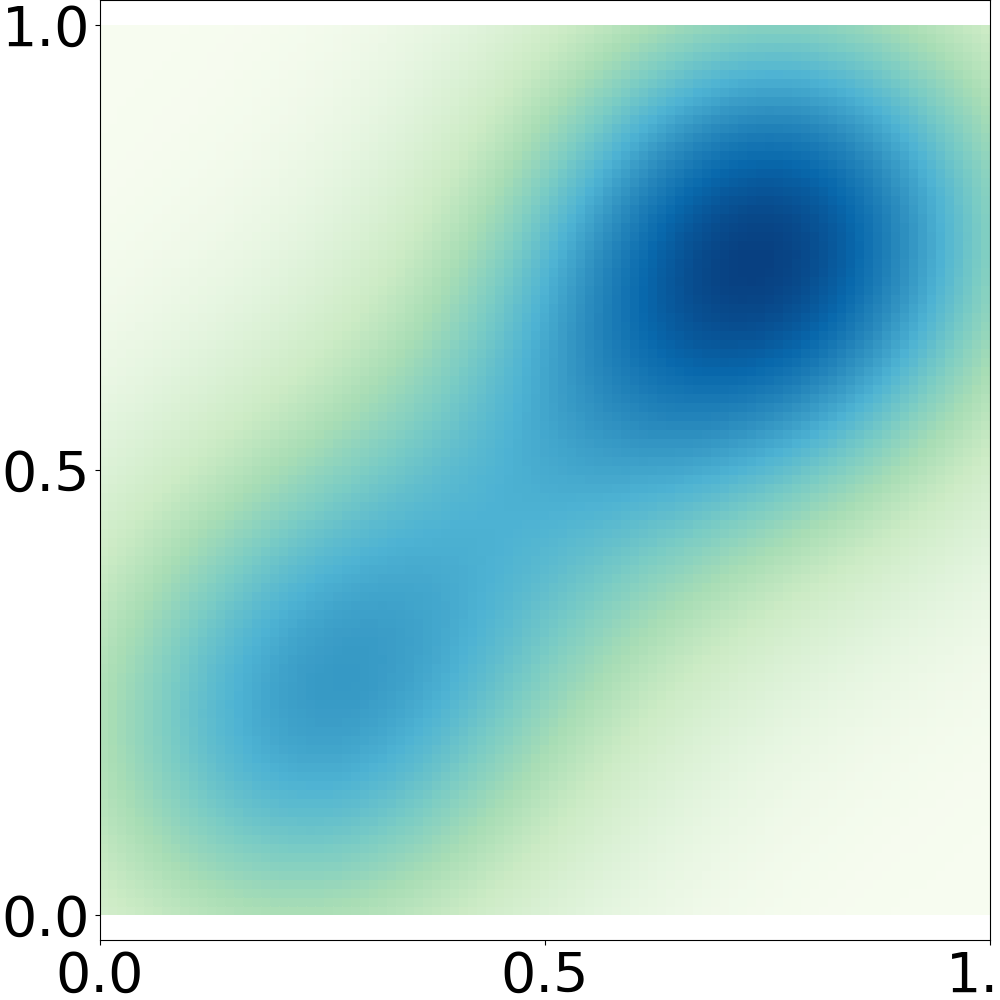}}\quad
\subfloat[VaGES-KSD ($L$=5)]{\includegraphics[width=0.19\columnwidth]{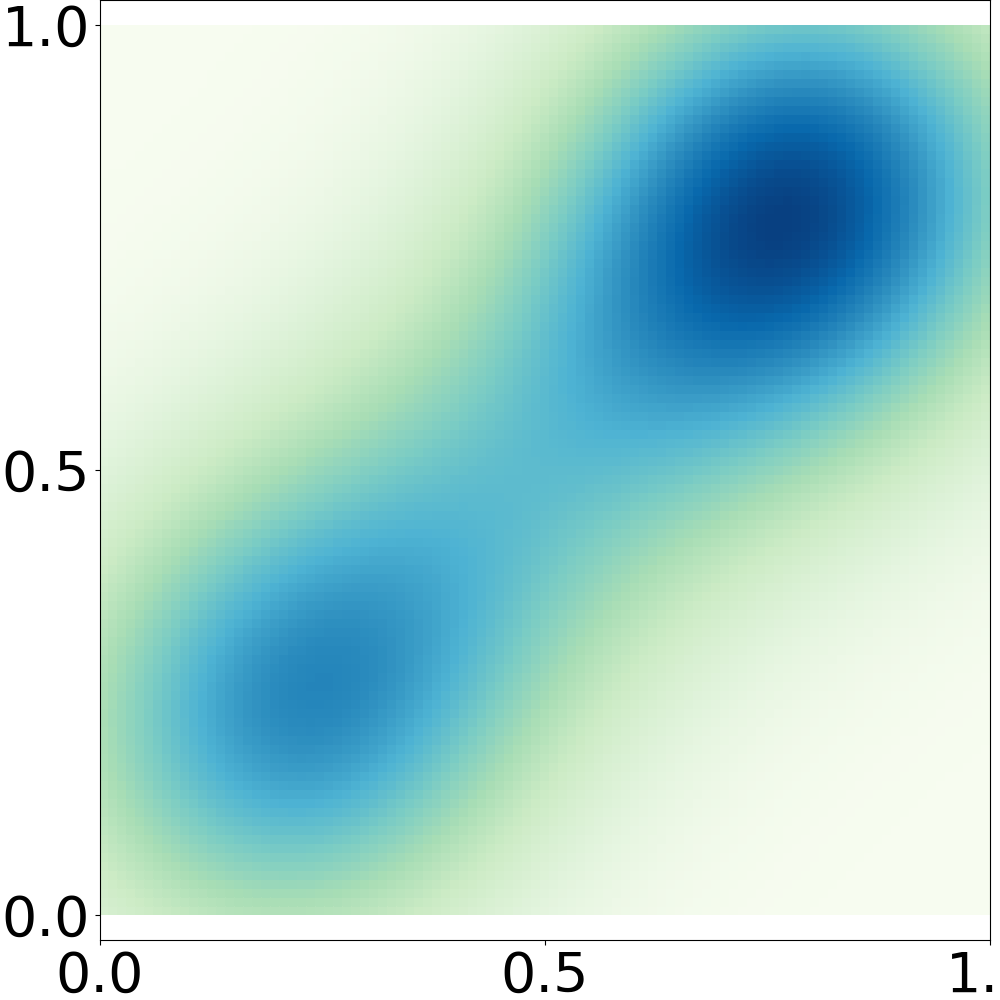}}\quad
\subfloat[VaGES-KSD ($L$=10)]{\includegraphics[width=0.19\columnwidth]{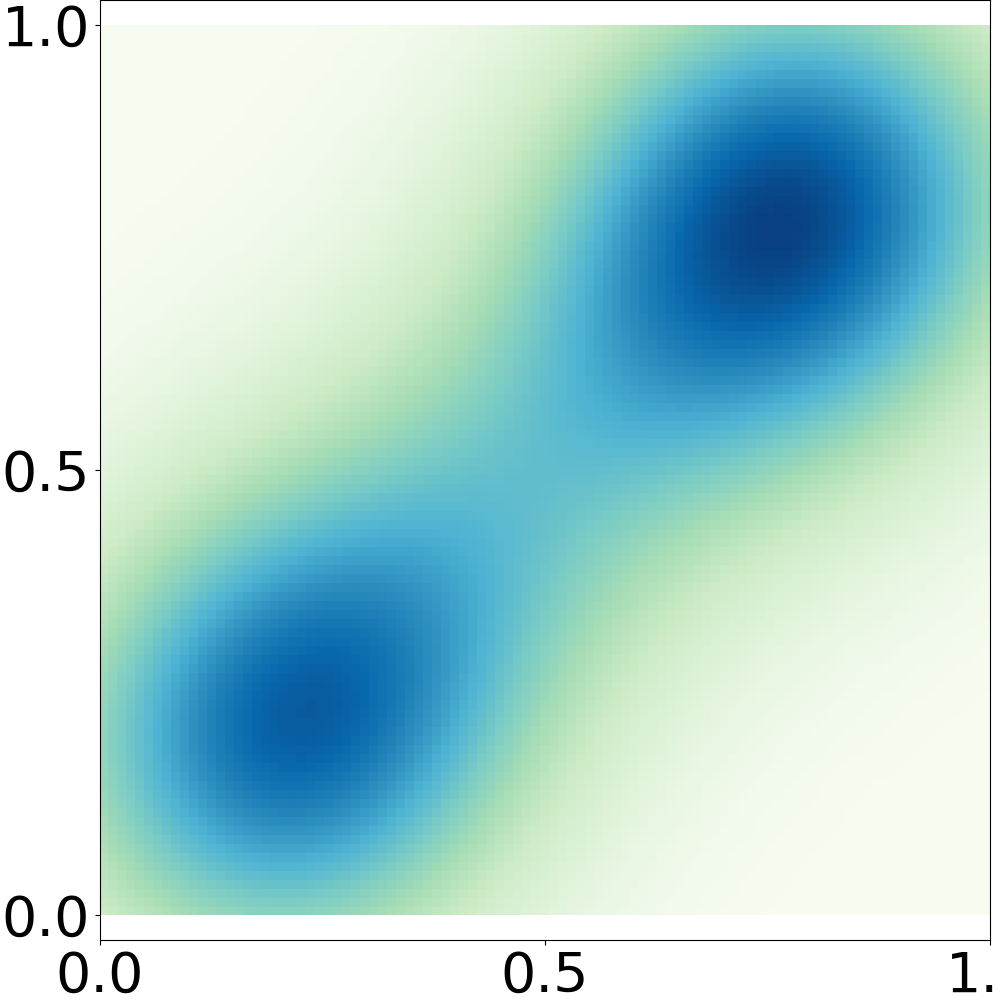}}\\
\vspace{-.2cm}
\subfloat[KSD]{\includegraphics[width=0.19\columnwidth]{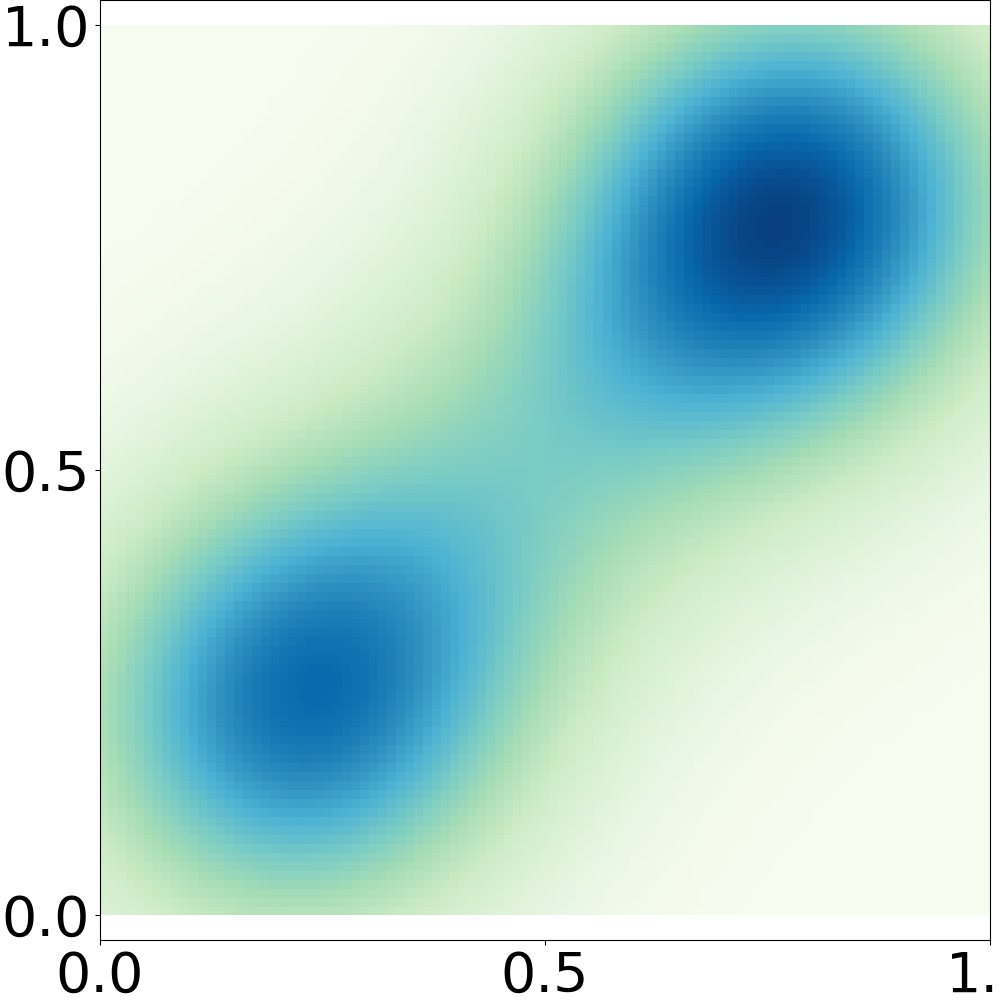}}\quad
\subfloat[IS-KSD ($L$=2)]{\includegraphics[width=0.19\columnwidth]{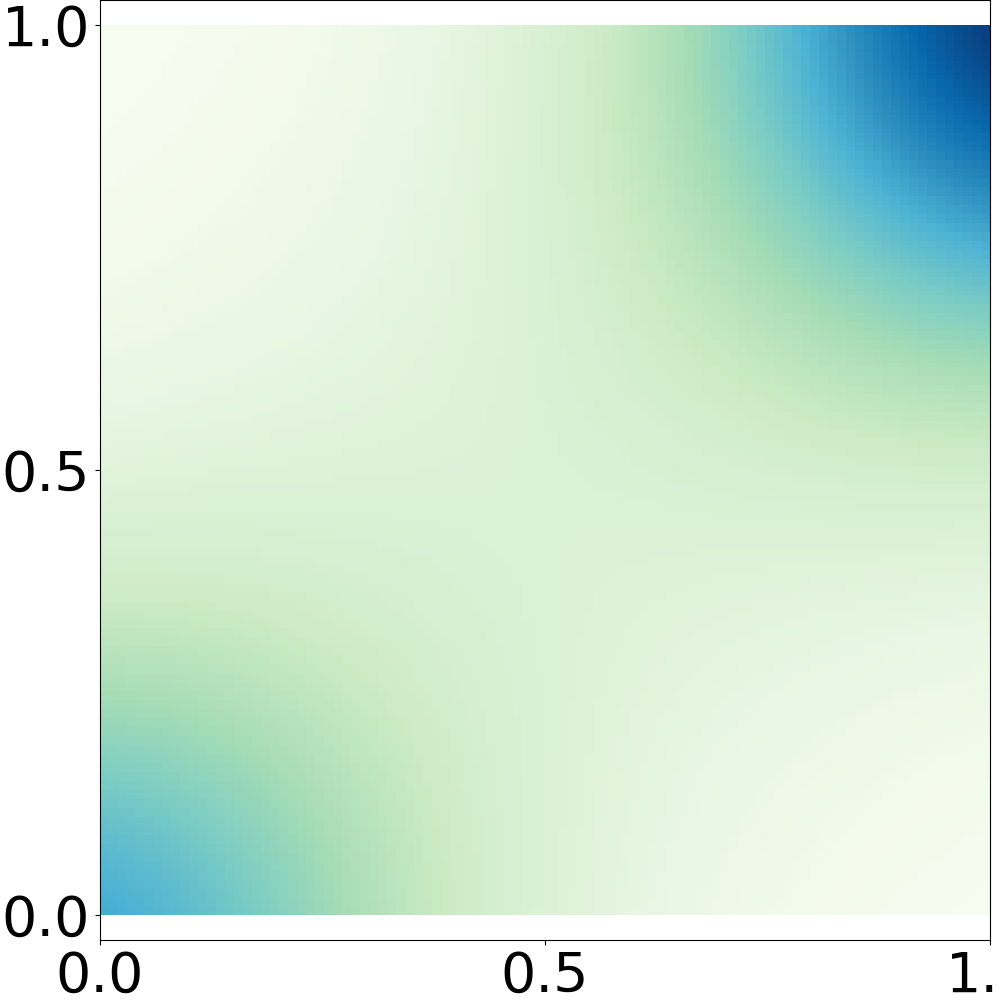}}\quad
\subfloat[IS-KSD ($L$=5)]{\includegraphics[width=0.19\columnwidth]{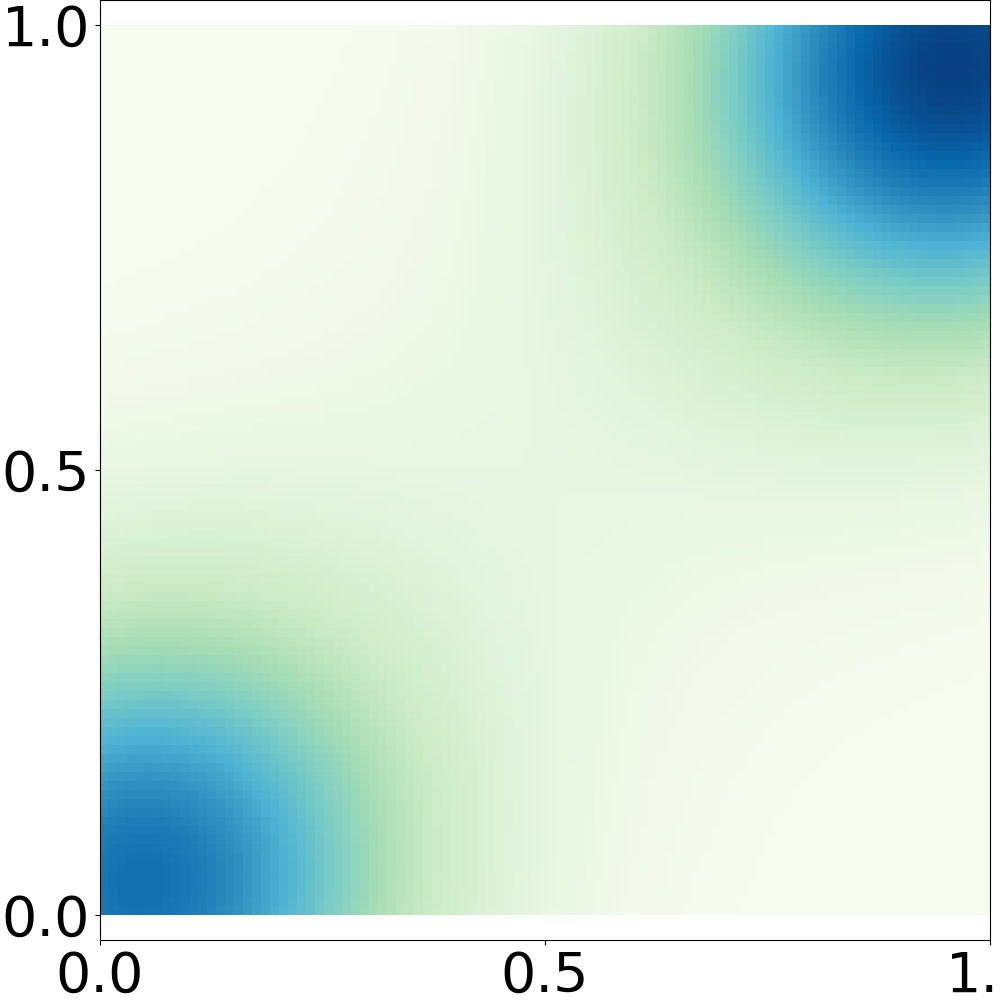}}\quad
\subfloat[IS-KSD ($L$=10)]{\includegraphics[width=0.19\columnwidth]{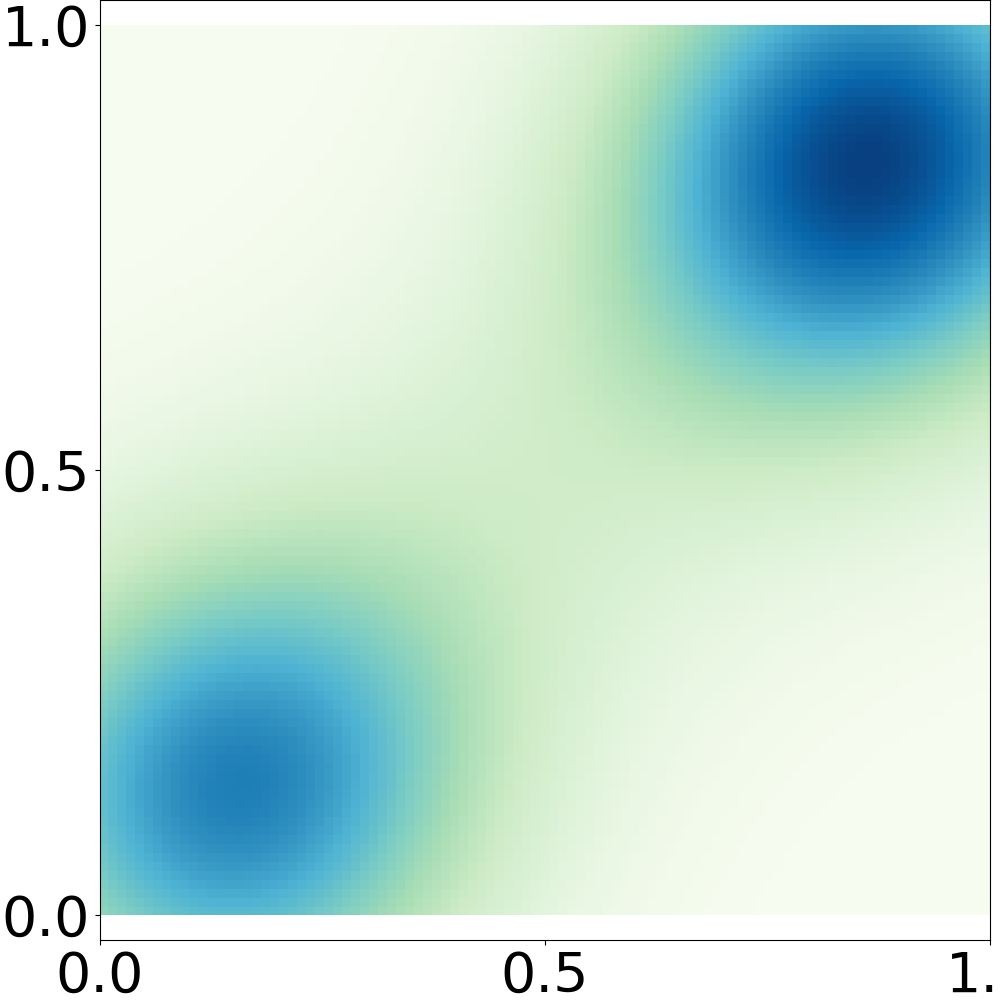}}\\
\caption{Density plots of GRBMs trained by KSD, VaGES-KSD and IS-KSD. $L$ is the number of samples from $q_\vphi(\vh|\vv)$.}
\label{fig:ksd_cmp}
\end{center}
\end{figure}

The density of the checkerboard dataset is shown in Fig.~\ref{fig:ksd_cmp} (a). The densities of GRBMs learned by KSD, VaGES-KSD and IS-KSD are shown in Fig.~\ref{fig:ksd_cmp} (b-h). Our VaGES-KSD is comparable to the KSD baseline and is better than the IS-KSD baseline. The result is consistent with the test log-likelihood results in Fig.~\ref{fig:ckbd_ksd} in the full paper.

\newpage

\subsection{Learning EBLVMs with Score Matching}
\subsubsection{Comparison in GRBMs}
\label{sec:add_cmp_grbm}

\begin{figure}[H]
    \centering
    \includegraphics[width=0.45\linewidth]{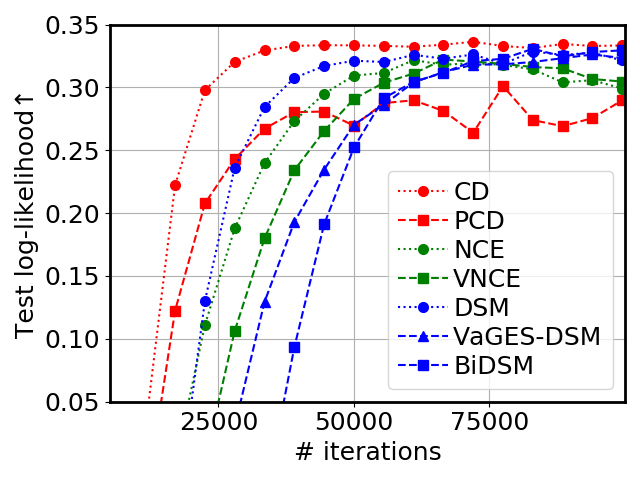}
    \caption{Comparison of different methods on checkerboard. The test log-likelihood is averaged over 10 runs.}
    \label{fig:ckbd_curve}
\end{figure}

\begin{figure}[H]
\begin{center}
\subfloat[Data density]{\includegraphics[width=0.19\columnwidth]{imgs/ksd/checkerboard.png}}\quad
\subfloat[CD]{\includegraphics[width=0.19\columnwidth]{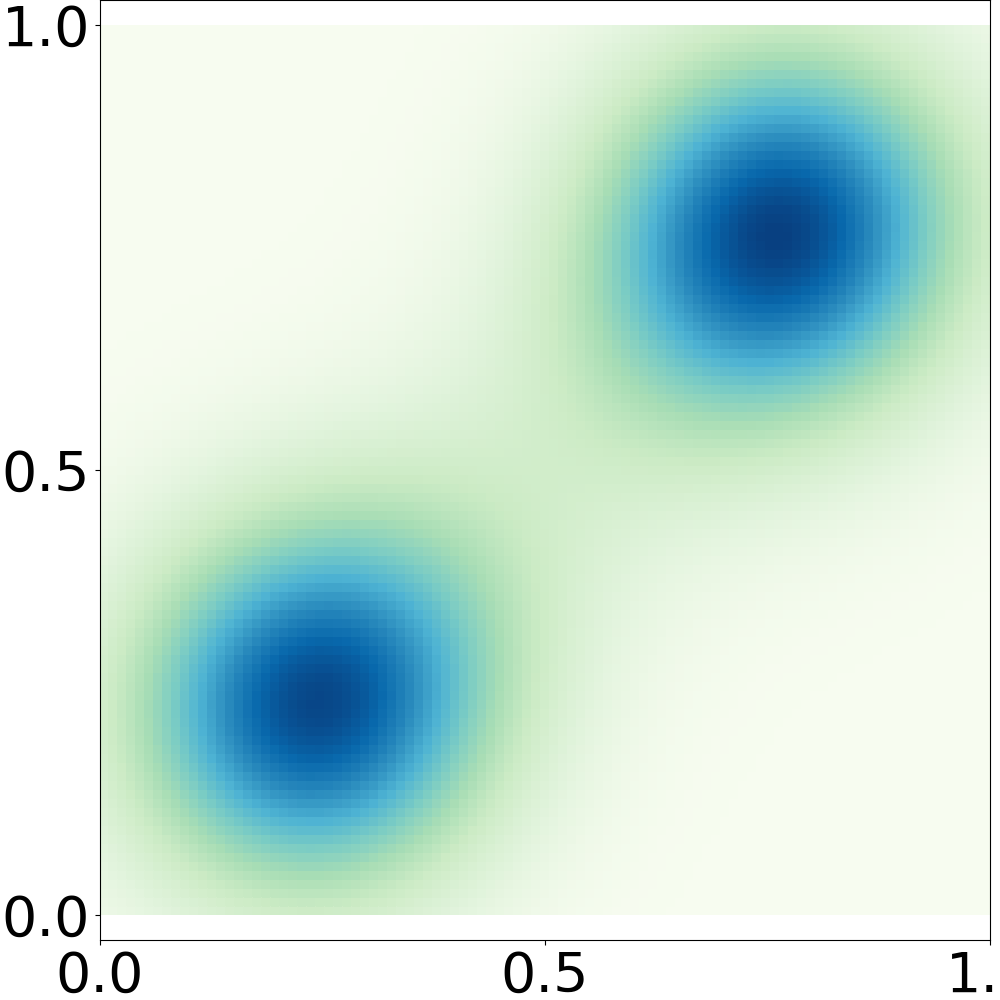}}\quad
\subfloat[PCD]{\includegraphics[width=0.19\columnwidth]{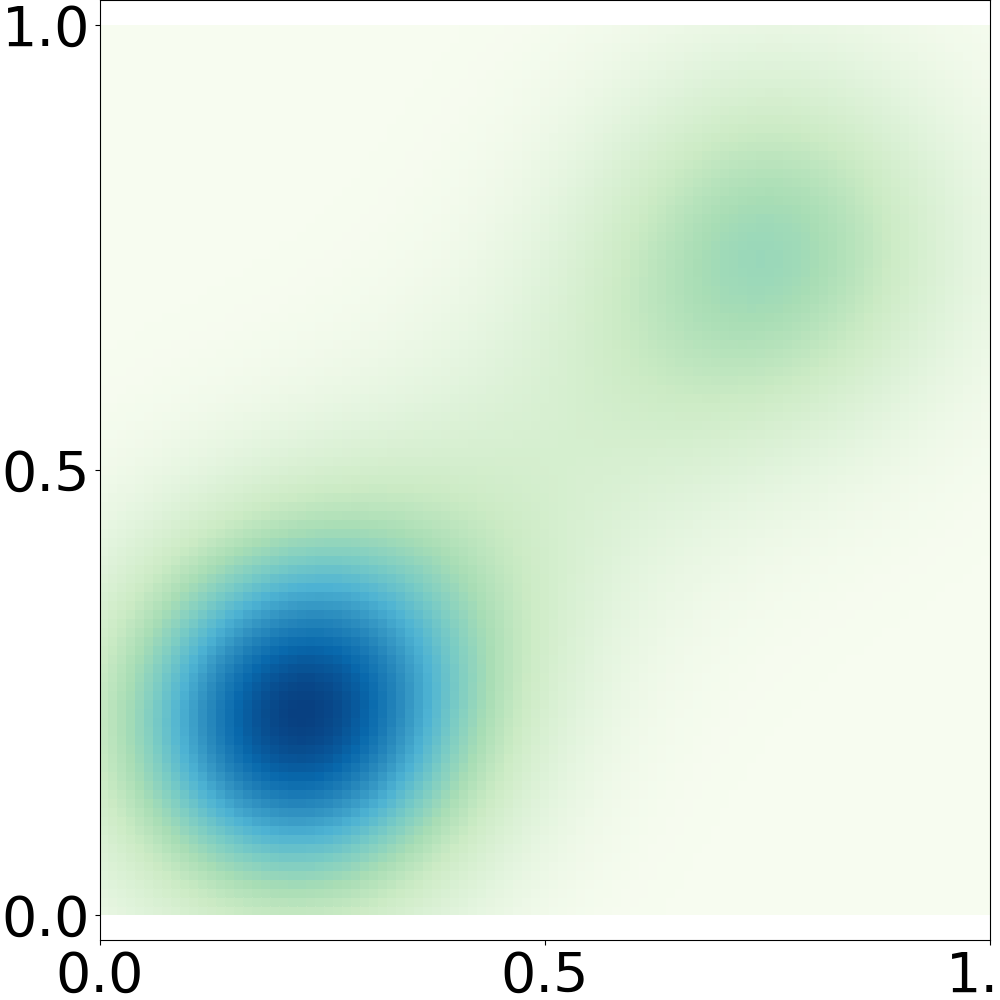}}\quad
\subfloat[NCE]{\includegraphics[width=0.19\columnwidth]{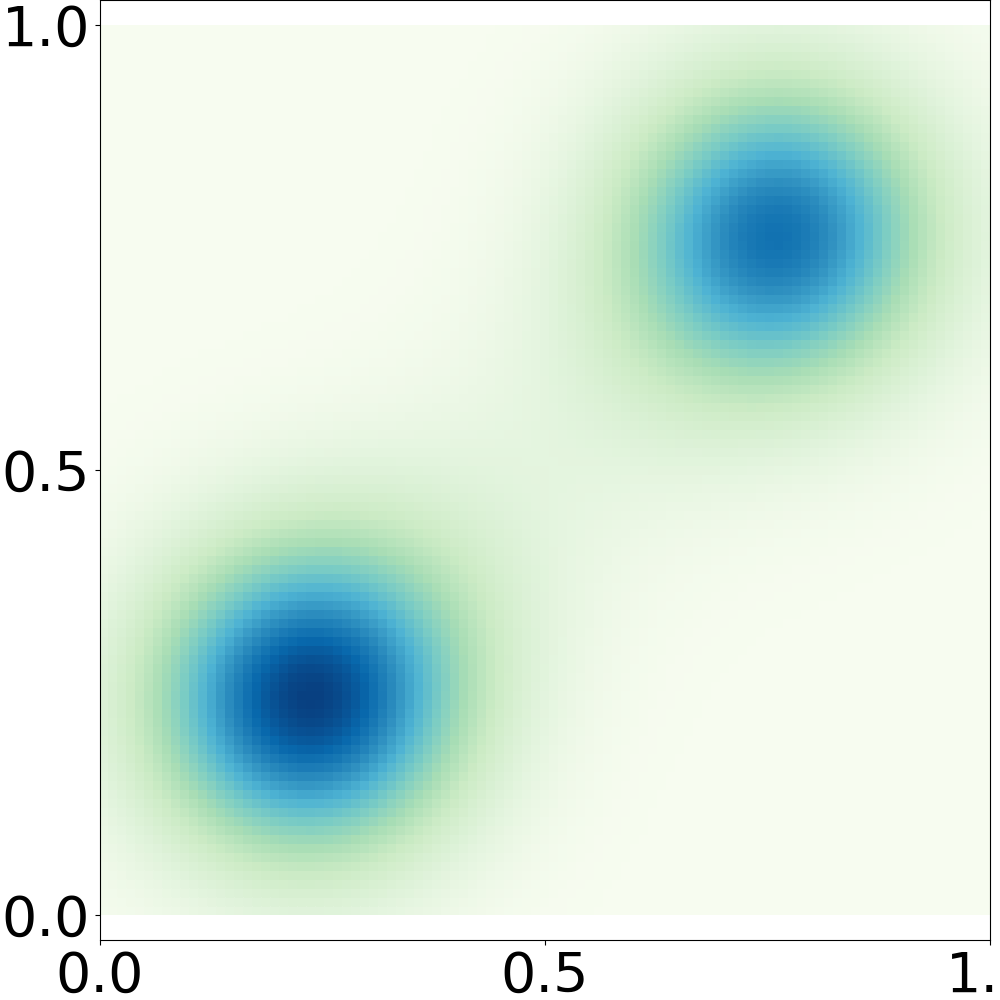}}\\
\vspace{-.2cm}
\subfloat[VNCE]{\includegraphics[width=0.19\columnwidth]{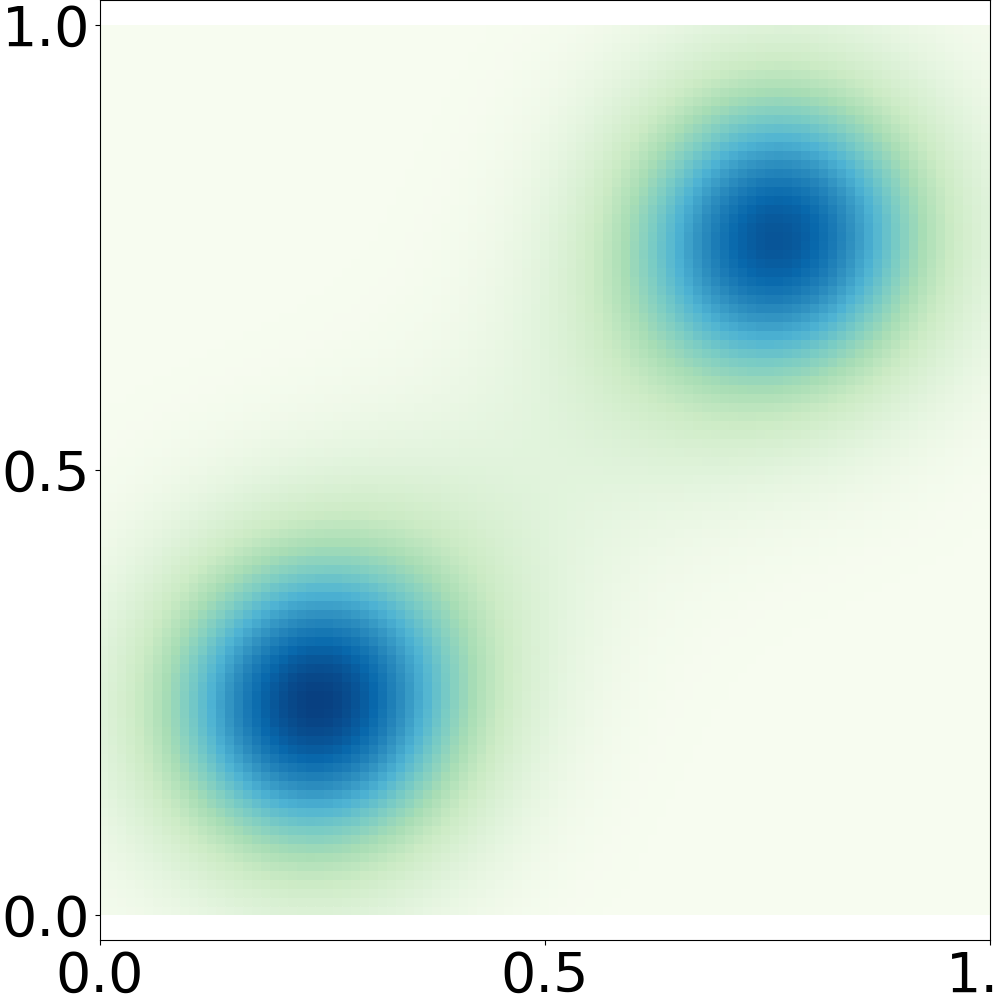}}\quad
\subfloat[DSM]{\includegraphics[width=0.19\columnwidth]{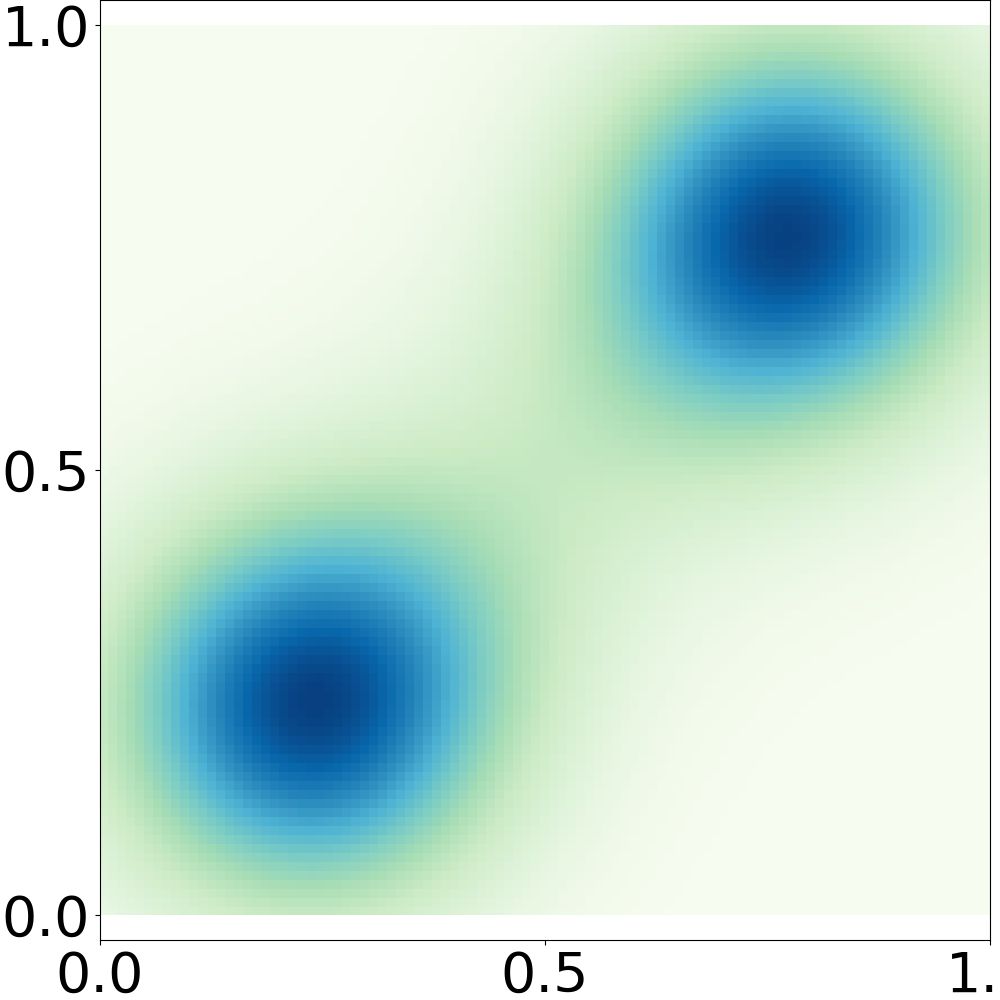}}\quad
\subfloat[VaGES-DSM]{\includegraphics[width=0.19\columnwidth]{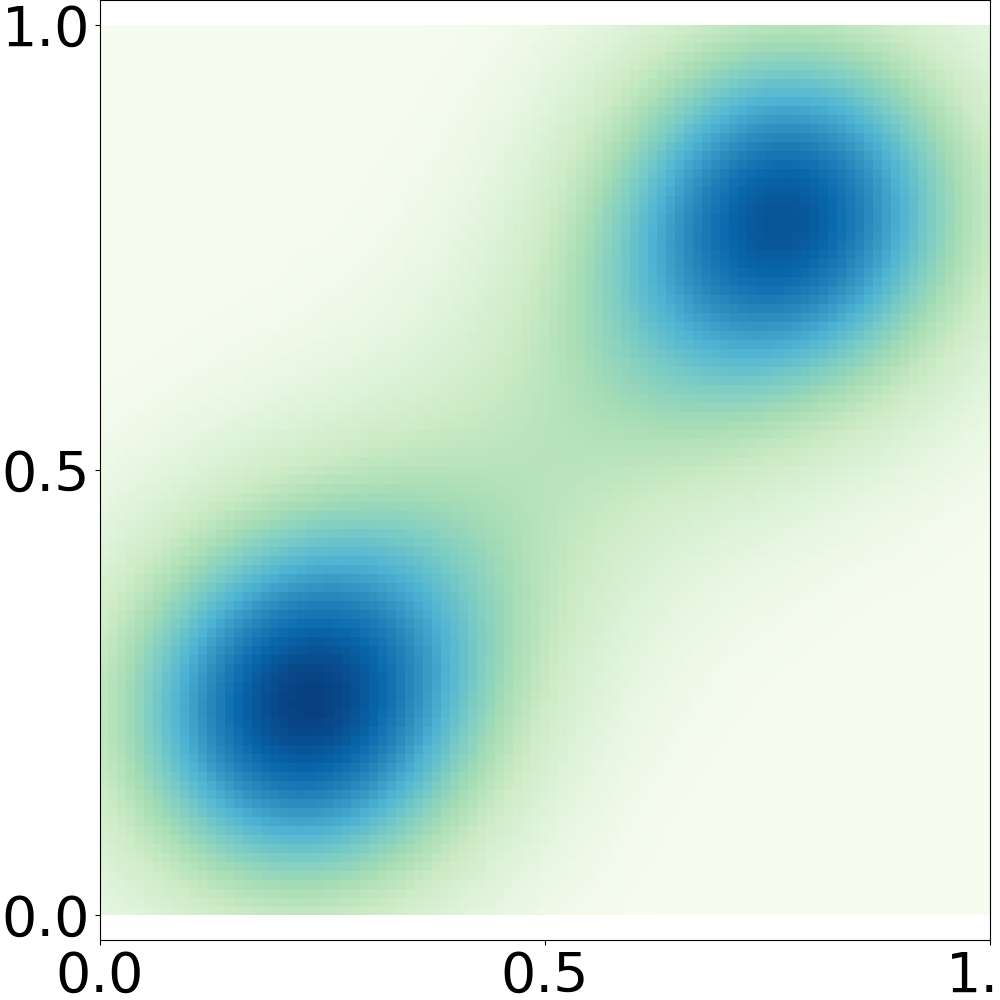}}\quad
\subfloat[BiDSM]{\includegraphics[width=0.19\columnwidth]{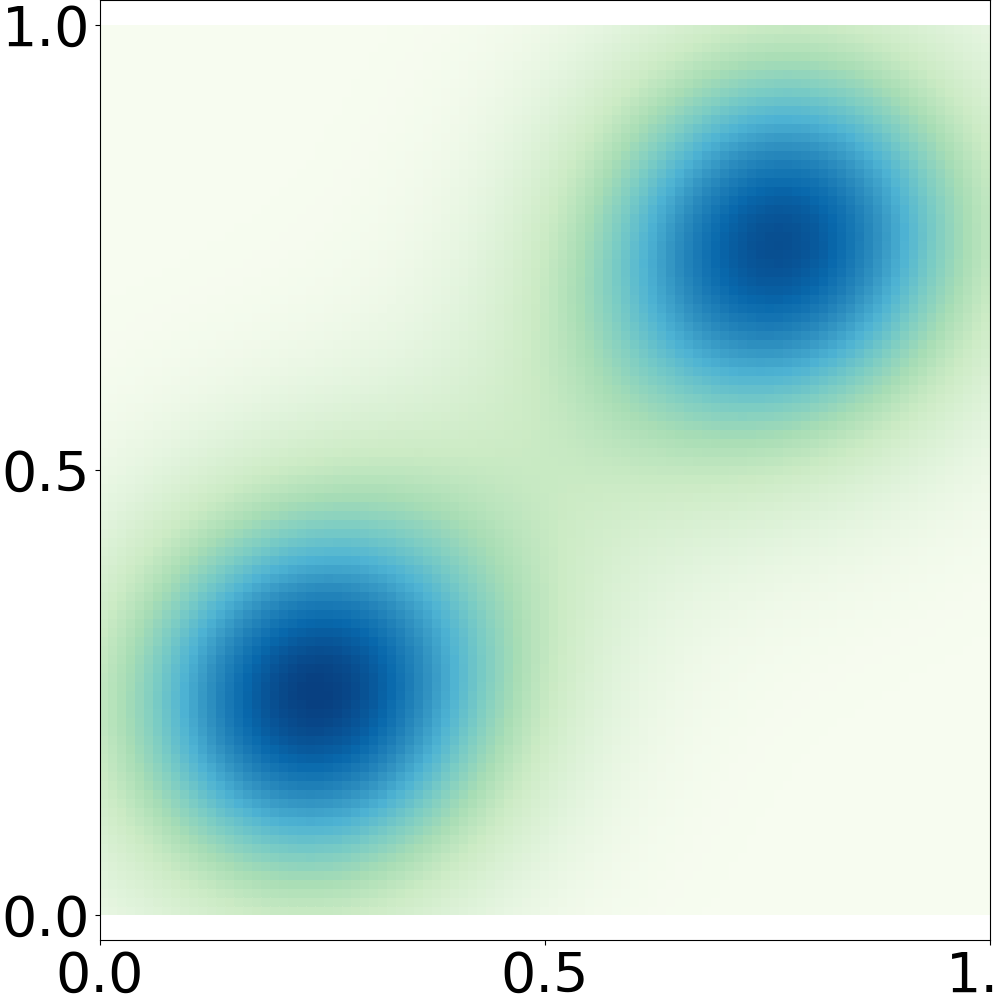}}\\
\caption{Density plots of GRBMs trained by different methods on checkerboard.}
\label{fig:sm_cmp}
\end{center}
\end{figure}

We compare with DSM~\cite{vincent2011connection}, BiSM~\citep{bao2020bilevel}, CD-based methods~\cite{hinton2002training,tieleman2008training} and noise contrastive estimation (NCE)-based methods~\cite{gutmann2010noise,pmlr-v89-rhodes19a} on the checkerboard dataset. In Fig.~\ref{fig:ckbd_curve}, we plot the test log-likelihood of different methods under the same setting. The result of VaGES-DSM is similar to CD, DSM and BiDSM and slightly better than PCD and NCE-based methods after convergence. The convergence speed of VaGES-DSM is faster than BiDSM. Besides, we show the densities of GRBMs learned by these methods in Fig.~\ref{fig:sm_cmp}. The performance of VaGES-DSM is similar to CD, DSM and BiDSM and better than PCD, NCE and VNCE, which agrees with the test log-likelihood results after convergence in Fig.~\ref{fig:ckbd_curve}.

\newpage
\subsubsection{Learning Deep EBLVMs}
\label{sec:add_learn_eblvm}
\begin{figure}[H]
\vspace{-.6cm}
\begin{center}
\subfloat[MNIST]{\includegraphics[width=0.4\columnwidth]{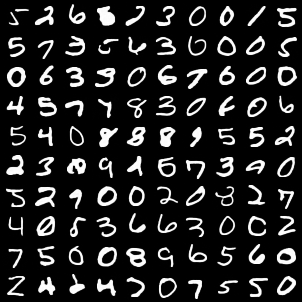}}\quad
\subfloat[CIFAR10]{\includegraphics[width=0.4\columnwidth]{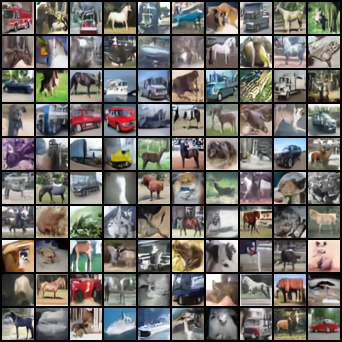}}\\
\vspace{-.3cm}
\subfloat[CelebA]{\includegraphics[width=0.75\linewidth]{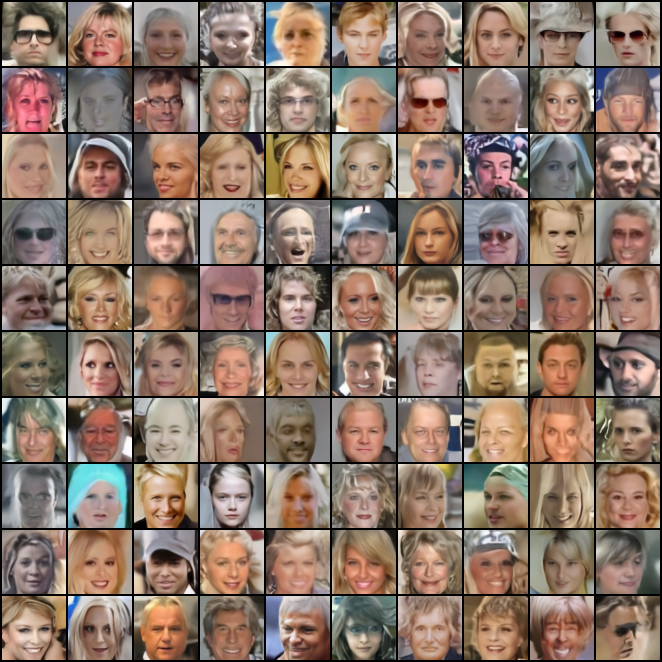}}\quad
\vspace{-.3cm}
\caption{Samples from EBLVMs.}
\label{fig:eblvm_sample}
\end{center}
\vspace{-.5cm}
\end{figure}

\vspace{-.1cm}
\textbf{Sample quality.} We show samples from EBLVMs learned on MNIST, CIFAR10 and CelebA in Fig.~\ref{fig:eblvm_sample}. We also evaluate the Inception Score on CIFAR10 and VaGES gets $7.53$, which is better than baselines such as VAE-EBLVM~\cite{han2020joint} ($7.17$) and EBM~\cite{du2019implicit} ($6.78$).

\begin{figure}[H]
\vspace{-0.4cm}
\begin{center}
\subfloat[MNIST]{\includegraphics[width=0.4\columnwidth]{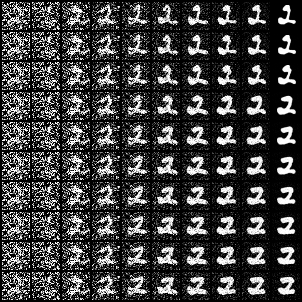}}\quad
\subfloat[CIFAR10]{\includegraphics[width=0.4\columnwidth]{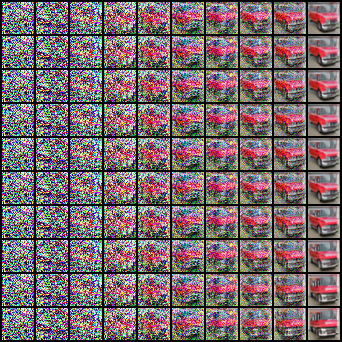}}\\
\vspace{-.1cm}
\subfloat[CelebA]{\includegraphics[width=0.75\linewidth]{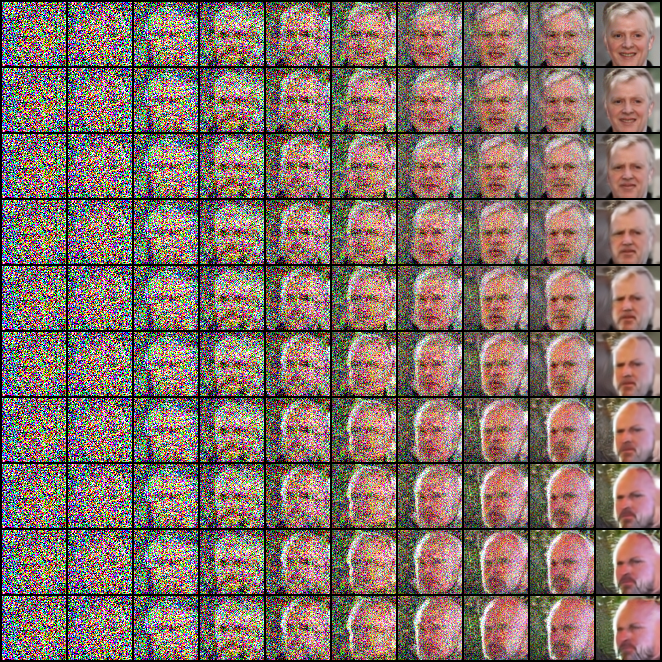}}\quad
\vspace{-.2cm}
\caption{Interpolation of annealed Langevin dynamics trajectories in the latent space in EBLVMs.}
\label{fig:eblvm_interpolation}
\end{center}
\vspace{-.5cm}
\end{figure}

\newpage
\textbf{Interpolation in the latent space.} We show more interpolation results in Fig.~\ref{fig:eblvm_interpolation}.

\newpage
\textbf{Sensitivity analysis.} We study how hyperparameters influence the performances of VaGES-SM in deep EBLVMs. The result is shown in Tab.~\ref{tab:sensitivity}. Increasing the number of convolutional layers will improve the performance, while the dimension of $\vh$, the number of $\vh$ sampled from $q_\vphi(\vh|\vv)$ and the noise level in Langevin dynamics don't affect the result very much. Setting both the number of times updating $\vphi$ and the number of Langevin dynamics steps to 5 is enough for a stable training. Besides, we also try using the KL divergence to learn the variational posterior and get a FID of 29.13, which doesn't affect the result very much.

\vspace{-.5cm}
\begin{table}[H]
\centering
\caption{\centering Sensitivity analysis on different hyperparameters (evaluated by FID $\downarrow$ on CIFAR10). Div means the training diverges.}
\label{tab:sensitivity}
\vspace{.3cm}
\hfill
\begin{minipage}[b]{.6\linewidth}
\subfloat[Dimensions of $\vh$]{\begin{tabular}{cccc}
\toprule
& 20 & 50 & 100\\
\midrule
FID & 26.55 & 28.09 & 27.78 \\
\bottomrule
\end{tabular}}\quad
\subfloat[\# convolutional layers]{\begin{tabular}{cccc}
\toprule
& 12 & 18 & 24\\
\midrule
FID & 36.17 & 28.09 & 25.98 \\
\bottomrule
\end{tabular}}\\
\subfloat[$\#$ $\vh$ sampled from $q_\vphi(\vh|\vv)$]{\begin{tabular}{cccc}
\toprule
& 2 & 5 & 10\\
\midrule
FID & 28.09 & 31.21 & 29.26 \\
\bottomrule
\end{tabular}}\quad
\subfloat[Noise level in Langevin dynamics]{\begin{tabular}{cccc}
\toprule
& $10^{-4}$ & $10^{-3}$ & $10^{-2}$\\
\midrule
FID & 28.09 & 25.83 & 30.47 \\
\bottomrule
\end{tabular}}
\end{minipage}%
\hspace{-1.cm}
\begin{minipage}[t]{.4\linewidth}
\vspace{-3.cm}
\captionsetup[subfloat]{captionskip=-31pt}
\subfloat[$\begin{array}{c}\\
\mbox{\# times updating }\vphi \mbox{ (}K\mbox{)}\\
\mbox{\# Langevin dynamics steps (}C\mbox{)}
\end{array}$]
{
\begin{adjustbox}{angle=-45}
\begin{tabular}{xx|xxxx}
& \multicolumn{1}{c}{} & \multicolumn{4}{c}{$C$} \\
 & \rotatebox[origin=c]{45}{FID} & 0 & 5 & 10 & 15\\
\cline{2-6}
\multirow{4}{*}[-1em]{\rotatebox[origin=c]{90}{$K$}}  & \rotatebox[origin=c]{90}{0} & \rotatebox[origin=c]{45}{Div} & \rotatebox[origin=c]{45}{Div} & \rotatebox[origin=c]{45}{Div} & \rotatebox[origin=c]{45}{Div}\\
& \rotatebox[origin=c]{90}{5} & \rotatebox[origin=c]{45}{Div} & \rotatebox[origin=c]{45}{28.09} & \rotatebox[origin=c]{45}{27.88}\\
& \rotatebox[origin=c]{90}{10} & \rotatebox[origin=c]{45}{Div} & \rotatebox[origin=c]{45}{29.52} & \\ 
& \rotatebox[origin=c]{90}{15} & \rotatebox[origin=c]{45}{Div} & \\
\end{tabular}
\end{adjustbox}}
\end{minipage}\hspace{-.3cm}
\end{table}
\vspace{-2.4cm}

\section{Additional Attempts on Improving Estimates}

\begin{figure}[H]
    \centering
    \includegraphics[width=0.45\linewidth]{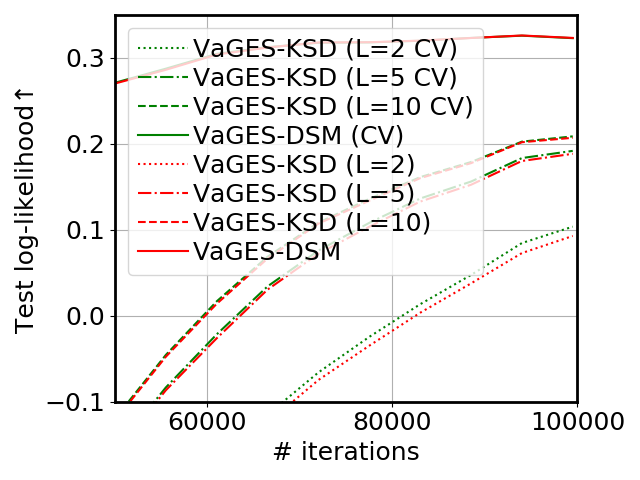}
    \vspace{-.4cm}
    \caption{How the control variate (CV) influences the performance over different objectives on the checkerboard dataset in GRBMs. The test log-likelihood is averaged over 10 runs.}
    \label{fig:if_vr}
\end{figure}

We can directly apply the control variate technique~\cite{mcbook} to VaES. By noticing that
\begin{align}
\label{eqn:zero_mean}
\E_{q_\vphi(\vh|\vv)} \nabla_\vv \log q_\vphi(\vh|\vv) = \int q_\vphi(\vh|\vv) \frac{\nabla_\vv q_\vphi(\vh|\vv)}{q_\vphi(\vh|\vv)} \mathrm{d}\vh = \nabla_\vv \int q_\vphi(\vh|\vv) \mathrm{d}\vh = \vzero,
\end{align}
we can subtract $\nabla_\vv \log q_\vphi(\vh|\vv)$ from VaES without changing the value of the expectation, and the resulting variational estimate is
\begin{align}
\label{eqn:control_variate}
    \text{VaES-CV}(\vv; \vtheta, \vphi) = \frac{1}{L} \sum\limits_{i=1}^L \nabla_\vv \log \frac{\tilde{p}_\vtheta(\vv, \vh_i)}{q_\vphi(\vh_i|\vv)},\quad \vh_i \stackrel{\text{i.i.d}}{\sim} q_\vphi(\vh|\vv).
\end{align}
When the variational posterior $q_\vphi(\vh|\vv)$ is equal to the true posterior $p_\vtheta(\vh|\vv)$, $\text{VaES-CV}(\vv; \vtheta, \vphi)=\nabla_\vv \log \frac{\tilde{p}_\vtheta(\vv, \vh)}{p_\vtheta(\vh|\vv)} = \nabla_\vv \log \tilde{p}_\vtheta(\vv)$ is exactly equal to the score function and has zero bias and variance. Empirically, we study how the control variate influences the performance over different objectives on the checkerboard dataset in GRBMs. As shown in Fig.~\ref{fig:if_vr}, the control variate only marginally improves the performance of VaGES-KSD and makes no difference to VaGES-DSM. As a result, we don't make the control variate a default technique in VaES, since it will introduce some extra computation, while the improvement is marginal.

\section{An Introduction to BiSM}
\label{sec:intro_bism}
BiSM~\cite{bao2020bilevel} approximates the score function via variational inference first:
\begin{align*}
\nabla_{\vv} \log p(\vv; \vtheta)  = \nabla_{\vv} \log  \frac{ \tilde{p}(\vv, \vh; \vtheta)}{p(\vh | \vv; \vtheta)} - \nabla_{\vv} \log \gZ(\vtheta)
= \nabla_{\vv} \log  \frac{ \tilde{p}(\vv, \vh; \vtheta)}{p(\vh | \vv; \vtheta)},
\end{align*}
and then gets the gradient of a certain objective via solving a complicated bi-level optimization problem:
\begin{align*}
\min_{\vtheta\in \Theta} \gJ_{Bi}(\vtheta, \vphi^*(\vtheta)), \;\; \gJ_{Bi}(\vtheta, \vphi) = \E_{q(\vv, \vepsilon)} \E_{q(\vh|\vv; \vphi)} \gF\left( \nabla_{\vv} \log  \frac{\tilde{p}(\vv, \vh; \vtheta)}{ q(\vh | \vv; \vphi)}, \vepsilon, \vv \right),
\end{align*}
where $\Theta$ is the hypothesis space of the model, $\gF$ depends on the certain objective, $q(\vv, \vepsilon)$ is the joint distribution of the data and additional noise and $\vphi^*(\vtheta)$ is defined as follows:
\begin{align*}
\vphi^*(\vtheta) = \mathop{\arg \min}\limits_{\vphi \in \Phi} \gG(\vtheta, \vphi), \text{ with } \gG(\vtheta, \vphi) = \E_{q(\vv, \vepsilon)} \gD \left(q(\vh | \vv; \vphi)|| p(\vh | \vv; \vtheta)\right).
\end{align*}
BiSM uses gradient unrolling to solve the problem, where the lower level problem $\vphi^*(\vtheta)$ is approximated by the output of $N$ steps gradient descent on $\gG(\vtheta, \vphi)$ w.r.t. $\vphi$, which is denoted by $\vphi^N(\vtheta)$. Finally, the model is updated with the approximate gradient $\nabla_\vtheta \gJ_{Bi}(\vtheta, \vphi^N(\vtheta))$, whose bias converges to zero in a linear rate in terms of $N$ when $G$ is strongly convex. The gradient unrolling requires an $O(N)$ time and memory.

Gradient unrolling of small steps is of large bias and that of large steps is time and memory consuming. Thus, BiSM with 0 gradient unrolling suffers from an additional bias besides the variational approximation. Instead, VaGES directly approximates the gradient of score function and its bias is controllable as presented in Sec.~\ref{sec:bd_bias}.

\end{document}